\documentclass[runningheads]{llncs}
\usepackage{graphicx}
\usepackage{color}

\usepackage{tikz}
\usepackage{comment}
\usepackage{subcaption}
\usepackage{algorithm}
\usepackage{algpseudocode}
\usepackage{booktabs}
\usepackage{xspace}
\usepackage{amsmath,amssymb} 
\usepackage{color}

\usepackage[accsupp]{axessibility}  

\usepackage[width=122mm,left=12mm,paperwidth=146mm,height=193mm,top=12mm,paperheight=217mm]{geometry}
\usepackage[pagebackref=true,breaklinks=true,letterpaper=true,colorlinks,bookmarks=false,citecolor=blue,linkcolor=blue]{hyperref}

\newcommand{\modelname}{\textsc{BLT}\xspace}

\newcommand{\bos}{\langle \text{bos} \rangle}
\newcommand{\eos}{\langle \text{eos} \rangle}

\makeatletter
\DeclareRobustCommand\onedot{\futurelet\@let@token\@onedot}
\def\@onedot{\ifx\@let@token.\else.\null\fi\xspace}

\def\eg{\emph{e.g}\onedot} 
\def\ie{\emph{i.e}\onedot} 
\def\cf{\emph{c.f}\onedot} 
\def\etc{\emph{etc}\onedot} 
 
\def\etal{\emph{et al}\onedot}
\makeatother

\definecolor{mygray}{gray}{0.4}
\newenvironment{mytiny}{\color{mygray}\tiny}{}
\usepackage[capitalize]{cleveref}
\crefname{section}{Sec.}{Secs.}
\Crefname{section}{Section}{Sections}
\Crefname{table}{Table}{Tables}
\crefname{table}{Tab.}{Tabs.}

\begin{document}
\pagestyle{headings}
\mainmatter
\def\ECCVSubNumber{7035}  

\title{BLT: Bidirectional Layout Transformer for Controllable Layout Generation} 


\titlerunning{BLT}
%
\author{Xiang Kong$^2$\thanks{Work done during their research internship at Google.}~, Lu Jiang$^{1}$, Huiwen Chang$^1$, Han Zhang$^1$, \\Yuan Hao$^1$, Haifeng Gong$^1$, Irfan Essa$^{1,3}$}
\authorrunning{X. Kong et al.}
\institute{$^1$ Google$^2$ LTI, Carnegie Mellon University, $^3$ Georgia Institute of Technology\\
\email{\tt\small xiangk@cs.cmu.edu, lujiang@google.com}
}

\maketitle

\begin{abstract}
Creating visual layouts is a critical step in graphic design. Automatic generation of such layouts is essential for scalable and diverse visual designs. To advance conditional layout generation, we introduce BLT, a bidirectional layout transformer. BLT differs from previous work on transformers in adopting non-autoregressive transformers. In training, BLT learns to predict the masked attributes by attending to surrounding attributes in two directions.
During inference, BLT first generates a draft layout from the input and then iteratively refines it into a high-quality layout by masking out low-confident attributes. The masks generated in both training and inference are controlled by a new hierarchical sampling policy. We verify the proposed model on six benchmarks of diverse design tasks. Experimental results demonstrate two benefits compared to the state-of-the-art layout transformer models. First, our model empowers layout transformers to fulfill controllable layout generation. Second, it achieves up to $10$x speedup in generating a layout at inference time than the layout transformer baseline. Code is released at \url{https://shawnkx.github.io/blt}.
\end{abstract}

\section{Introduction}\label{sec:intro}

Graphic layout dictates the placement and sizing of graphic components, playing a central role in how viewers interact with the information provided~\cite{lee2020neural}. Layout generation is emerging as a new research area with a focus of generating realistic and diverse layouts to facilitate design tasks. Recent works show promising progress for various applications such as graphic user interfaces~\cite{deka2017rico,jiang2022coarse}, presentation slides~\cite{guo2021layout}, magazines~\cite{zheng2019content,yamaguchi2021canvasvae}, scientific publications~\cite{arroyo2021variational}, commercial advertisements~\cite{lee2020neural,qian2020retrieve,guo2021vinci}, computer-aided design~\cite{willis2021engineering}, indoor scenes~\cite{di2021multi}, layout representations~\cite{manandhar2020learning,xie2021canvasemb}, \etc.

\begin{figure}
\centering
\begin{subfigure}[b]{0.75\textwidth}
   \centering
   \includegraphics[width=\linewidth]{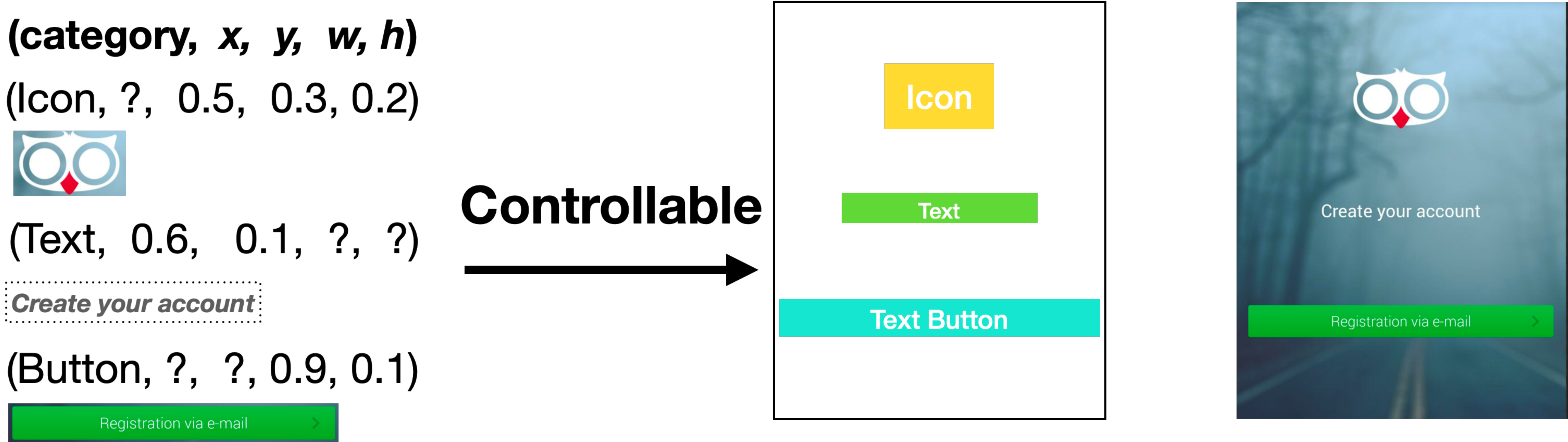}
   \caption{Conditional layout generation.}
   \label{fig:teaser} 
\end{subfigure}
\begin{subfigure}[b]{0.75\textwidth}
    \centering
   \includegraphics[width=1\linewidth]{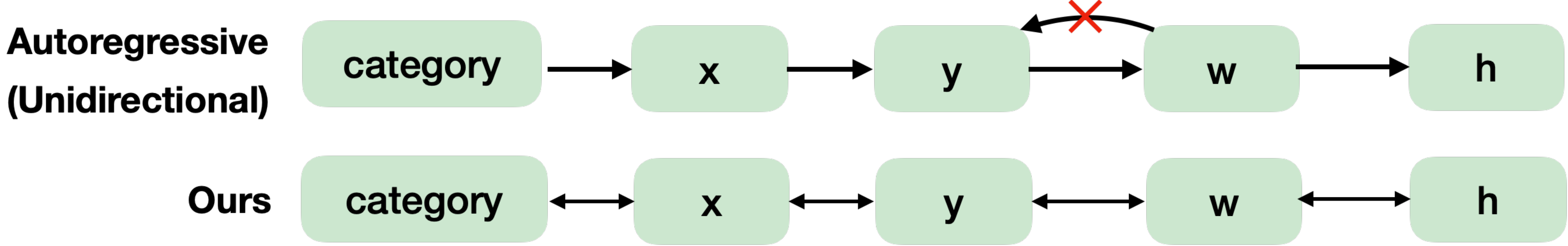}
   \caption{Unidirectional autoregressive (top) and non-autoregressive (bottom) decoding.}
   \label{fig:bidir}
\end{subfigure}
\caption{\textbf{(a) Conditional layout generation}. Each object is modeled by 5 attributes `category', `$x$', `$y$', `$w$' (width) and `$h$' (height). In conditional generation, attributes are partially given by the user and the goal is to generate the unknown attributes, \eg putting the icon or button on the canvas. \textbf{(b) Illustration of immutable dependency chain in autoregressive decoding.}}
\end{figure}

Previous work explores neural models for layout generation using Generative Adversarial Networks (GANs)~\cite{goodfellow2014generative,li2019layoutgan} or Variational Autoencoder (VAEs)~\cite{kingma2013auto,jyothi2019layoutvae,patil2020read,lee2020neural}. Currently, layout transformers hold the state-of-the-art performance for layout generation~\cite{gupta2020layout,arroyo2021variational}.
These transformers represent a layout as a sequence of objects and an object as a (sub)sequence of attributes (See Fig.~\ref{fig:teaser}).
Layout transformers predict the attribute sequentially based on previously generated output (\ie autoregressive decoding). Like other vision tasks, by virtue of the powerful self-attention~\cite{vaswani2017attention}, 
transformer models yield superior quality and diversity than GAN or VAE models for layout generation~\cite{gupta2020layout,arroyo2021variational}.

Unlike Layout VAE (or GAN) models that are capable of generating layouts considering user requirements, layout transformers, however, have difficulties in conditional generation as a result of an acknowledged limitation discussed in~\cite{gupta2020layout} (\cf order of primitives). Fig.~\ref{fig:teaser} illustrates a scenario in which a designer has objects with partially known attributes and hopes to generate the missing attributes. Specifically, each object is modeled by five attributes `category', `$x$', `$y$', `$w$' (width) and `$h$' (height). The designer wants the layout model to 1) place the ``icon'' and ``button'' with known sizes onto the canvas (\ie generating $x$, $y$ from $w$, $h$, and `category'), and 2) determines the size of the centered ``text object'' (\ie generating $w$, $h$ from $x$, $y$, and `category').

Such functionality is currently missing in the layout transformers~\cite{gupta2020layout,arroyo2021variational} due to \emph{immutable dependency chain}. This is because autoregressive transformers follow a pre-defined generation order of object attributes. As shown in Fig.~\ref{fig:bidir}, attributes must be generated starting from the category $c$, then $x$ and $y$, followed by $w$ and $h$. The dependency chain is immutable \ie it cannot be changed at decoding time. Therefore, autoregressive transformers fail to perform conditional layout generation when the condition disagrees with the pre-defined dependency, \eg generating position $y$ from the known width $w$ in Fig.~\ref{fig:bidir}.

In this work, we introduce Bidirectional Layout Transformer (or \modelname) for controllable layout generation. Different from the traditional transformer models~\cite{gupta2020layout,arroyo2021variational}, \modelname enables controllable layout generation where every attribute in the layout can be  modified, with high flexibility, based on the user inputs (\cf Fig.~\ref{fig:teaser}).
During training, \modelname learns to predict the masked attributes by attending to attributes in two directions (\cf Fig.~\ref{fig:blt_train}). At inference time, \modelname adopts a non-autoregressive decoding algorithm to refine the low-confident attributes iteratively into a high-quality layout (\cf Fig.~\ref{fig:blt_decode}). We propose a simple hierarchical sampling policy that is used both in training and inference to guide the mask generation over attribute groups.

\modelname eliminates a critical limitation in the prior layout transformer models~\cite{gupta2020layout,arroyo2021variational} that prevents transformers from performing controllable layout generation.
Our model is inspired by the autoregressive work in NLP~\cite{ghazvininejad2019mask,gu2017non,donahue2020end,gu-kong-2021-fully}. However, we find directly applying the non-autoregressive translation models~\cite{ghazvininejad2019mask,kong2020incorporating} to layout generation only leads to inferior results than the autoregressive baseline. 
Our novelty lies in the proposed simple yet novel hierarchical sampling policy, which, as substantiated by our experiments in Section~\ref{sec:order}, is essential for high-quality layout generation.

We evaluate the proposed method on six layout datasets under various metrics. 
These datasets cover representative design applications for graphic user interface~\cite{deka2017rico}, magazines~\cite{zheng2019content} and publications~\cite{zhong2019publaynet}, commercial ads~\cite{lee2020neural}, natural scenes~\cite{lin2014microsoft} and home decoration~\cite{fu20203dfront}. Experiments demonstrate two benefits to several strong baseline models~\cite{gupta2020layout,arroyo2021variational,jyothi2019layoutvae,jyothi2019layoutvae}. First, our model empowers transformers to fulfill controllable layout generation and thereby outperforms the previous conditional models based on VAE (\ie, LayoutVAE~\cite{jyothi2019layoutvae} and NDN~\cite{lee2020neural}).
Even though our model is not designed for unconditional layout generation, it achieves quality on-par with the state-of-the-art. Second, our new method reduces the time complexity in~\cite{gupta2020layout,arroyo2021variational} while achieving 4x-10x speedups in layout generation.

To summarize, we make the following contributions:
\begin{enumerate}
    \item We address a critical limitation in state-of-the-art layout transformers~\cite{gupta2020layout,arroyo2021variational} and hence empower transformers to fulfill controllable layout generation.
    \item Though our idea is inspired by the autoregressive work in NLP~\cite{ghazvininejad2019mask,gu2017non,donahue2020end,gu-kong-2021-fully}, a novel hierarchical mask sampling policy is introduced in training and decoding, which is essential for high-quality layout generation.
    \item Extensive experiments validate that our method performs favorably against state-of-the-art models in terms of realism, alignment, and semantic relevance on six diverse layout benchmarks.
\end{enumerate}

\section{Related Work}
\noindent\textit{Layout synthesis:}
Recently, automatic generation of high-quality and realistic layouts has fueled increasing interest. Unlike early work~\cite{o2014learning,o2014exploratory,pang2016directing,qi2018human,yu2011make,merrell2011interactive,fisher2012example,wang2018deep,fan2017point,wang2019planit}, recent data-driven methods rely on deep generative models such as GANs~\cite{goodfellow2014generative} and VAE~\cite{kingma2013auto}. For example, LayoutGAN~\cite{li2019layoutgan} uses a GANs-based framework to synthesize semantic and geometric properties for scene elements. During inference time, LayoutGAN generates layouts from the Gaussian noise. Afterwards, LayoutGAN is extended to attribute-conditioned design tasks~\cite{li2020attribute}. LayoutVAE~\cite{jyothi2019layoutvae} introduces two conditional VAEs. The first aims to learn the distribution of category counts which will be used during layout generation. The second produces layouts conditioning on the number and category of objects generated from the first VAE or ground-truth data. Recently, various VAE models are proposed~\cite{patil2020read,kikuchi2021constrained,lee2020neural}. Among them, Neural Design Networks (NDN)~\cite{lee2020neural} is a competitive VAEs-based model for conditional layout generation, which focuses on modeling the asset relations and constraints by graph convolution. Our work is different from LayoutVAE and NDN in modeling layout and user inputs by the transformer, which, as shown in Table~\ref{tab:cond_rst}, perform more favorably thanks to the transformer architecture. Our finding is consistent with \cite{arroyo2021variational} where Arroyo~\etal find VAEs underperforming transformers for unconditional layout generation~\cite{arroyo2021variational}.

Currently, the state-of-the-art for layout generation is held by the transformer models~\cite{vaswani2017attention}. In particular, \cite{gupta2020layout} employs the standard autoregressive Transformer decoder with unidirectional attention. They find out that self-attention is able to explicitly learn relationships between objects in the layout, resulting in superior quality compared to prior work. Furthermore, to increase the diversity of generated layout, \cite{arroyo2021variational} incorporates the standard autoregressive Transformer decoder into a VAE framework and \cite{nguyen2021diverse} employs multi-choice prediction and winner-takes-all loss. Despite the superior performance, this work addresses a critical limitation acknowledged in~\cite{gupta2020layout} that prevents transformers from performing controllable layout generation. Following LayoutGAN~\cite{li2019layoutgan}, \cite{kikuchi2021constrained} proposes a
Transformer based layout GAN model, LayoutGAN++. In this framework, the input is a set of asset labels and randomly generated code and the output is the location and size of these asset.Different from the LayoutGAN++, the input to our proposed model is more flexible and can support unconditional generation and various types of conditional generation tasks.

\noindent\textit{Bidirectional transformer and non-autoregressive decoding:}
The classic Transformer~\cite{vaswani2017attention} decoder uses the unidirectional self-attention mechanism to generate the sequence token-by-token from left to right, leaving the right-to-left contexts unexploited. Several NLP works~\cite{ghazvininejad2019mask,kong2020incorporating,stern2019insertion} are proposed to investigate language generation tasks by non-autoregressive generation with bidirectional Transformers, which allow representations to attend in both directions~\cite{devlin2019bert}.
However, non-autoregressive decoding process leads to an apparent performance drop compared to the autoregressive decoding algorithm~\cite{gu2017non,gu-kong-2021-fully,ghazvininejad2019mask}.
In this work, we finds that applying the non-autoregressive NLP model~\cite{ghazvininejad2019mask} to layout generation also leads to inferior results than the autoregressive baseline. To this end, we propose a simple yet effective hierarchical sampling policy which is essential for high-quality layout generation.

\section{Problem Formulation}

Following~\cite{gupta2020layout}, we use 5 attributes to describe an object, \ie, ($c$, $x$, $y$, $w$, $h$), in which the first element $c\in C$ is the object category such as the logo or button, and the remainder details the bounding box information \ie the center location $(x, y)\in\mathbb{R}^{2}$ and the width and height $(w, h)\in \mathbb{R}^{2}$. Furthermore, float values in bounding box information is discretized using 8-bit uniform quantization. For instance, the $x$-coordinate after the quantization becomes $\{x | x \in \mathbb{Z}, 0 \le x \le 31 \}$.
A layout $l$ of $K$ assets is hence denoted as a flattened sequence of integer indices:
\begin{align}
\label{eq:seq_def}
l = [\bos, c_{1}, x_{1}, y_{1}, w_{1}, h_{1}, c_{2}\cdots, h_{K}, \eos]
\end{align}
where $\bos$ and $\eos$ are special tokens to denote the start and the end of sequence. We use a shared vocabulary and represent each element in $l$ as an integer index or equivalently as a one-hot vector with the same length. It is trivial to extend the attribute dimension to model more complex layouts.

\paragraph{Issues} To train the model, prior work~\cite{gupta2020layout,arroyo2021variational} estimates the joint likelihood of observing a layout as $p(l) = \prod_{i=1}^{|l|}p(l_{i}|l_{1:i})$.

During training, an autoregressive Transformer model is learned to maximize the likelihood using ground-truth attribute as input (\ie teacher forcing). At inference time, the transformer model predicts the attribute sequentially based on previously generated output (\ie autoregressive decoding), starting from the begin-of-sequence or  $\bos$ token until yielding the end-of-sequence token $\eos$. The generation must follow a fixed conditional dependency. For example, Eq.~\eqref{eq:seq_def} defines an immutable generation order $x \rightarrow y \rightarrow w \rightarrow h$. And in order to generate the height $h$ for an object, one must know its $x$-$y$ coordinates and width $w$.

There are two issues with autoregressive decoding for the conditional generation. First, it is infeasible to process user conditions that differ from the dependency order used in training. For instance, the model using Eq.~\eqref{eq:seq_def} is not able to generate $x$-$y$ coordinates from width and height, which corresponds to a practical example of placing an object with given size.
This issue is exacerbated by complex layouts that require more attributes to represent an object. 
Second, the autoregressive inference is not parallelizable, rendering it inefficient for the dense layout with a large number of objects or attributes.

\section{Approach}\label{sec:blg}
Our goal is to design a transformer model for controllable layout generation.
We propose a method to learn non-autoregressive transformers.
Unlike existing layout transformers~\cite{gupta2020layout,arroyo2021variational}, the new layout transformer is bidirectional and can generate all attributes simultaneously in parallel, which allows not only for flexible conditional generation but also more efficient inference.
In this section, we first discuss the model and training objective; then detail a novel hierarchical sampling policy for training and parallel decoding.

\subsection{Model and Training}\label{sec:training}
\begin{figure*}[ht]
\centering
\begin{subfigure}[b]{0.35\textwidth}
   \includegraphics[width=\textwidth]{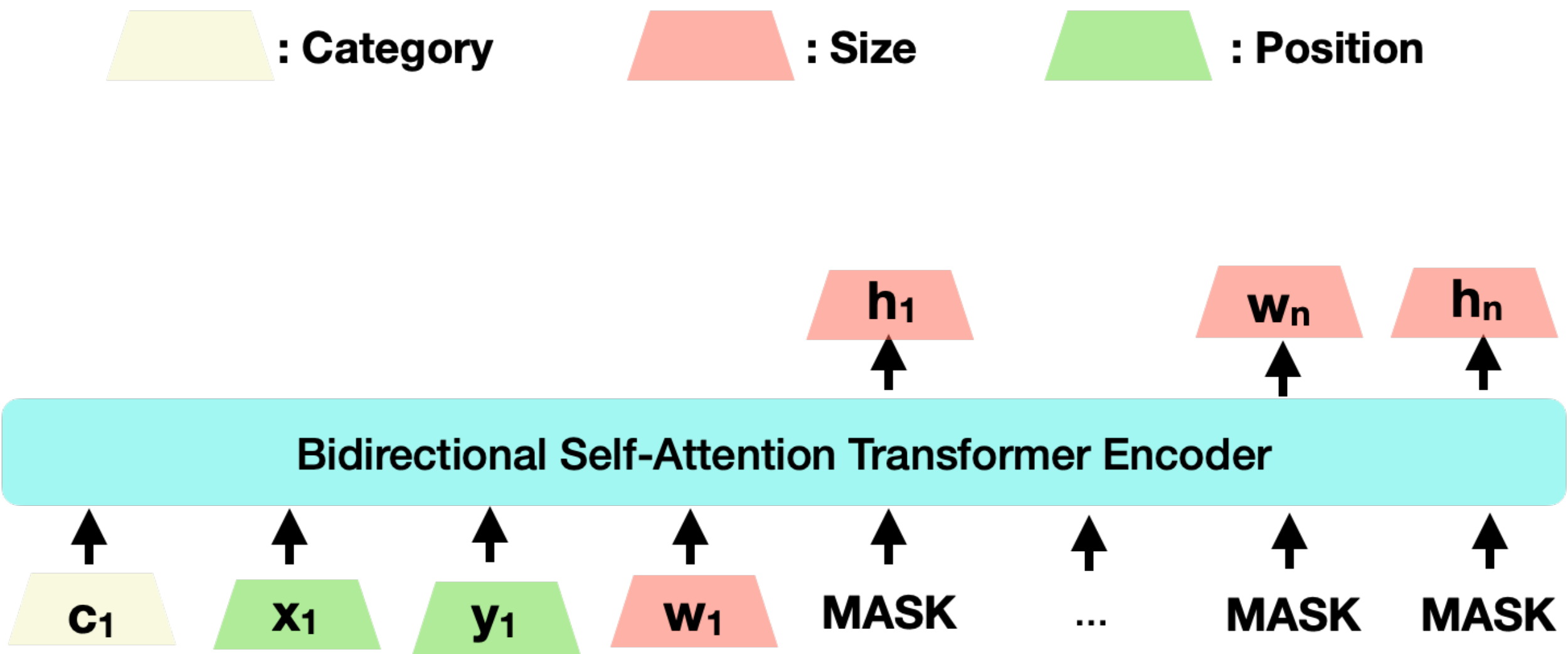}
   \caption{BLT Training Phrase.}
   \label{fig:blt_train} 
\end{subfigure}
~
\begin{subfigure}[b]{0.6\textwidth}
   \includegraphics[width=1\textwidth]{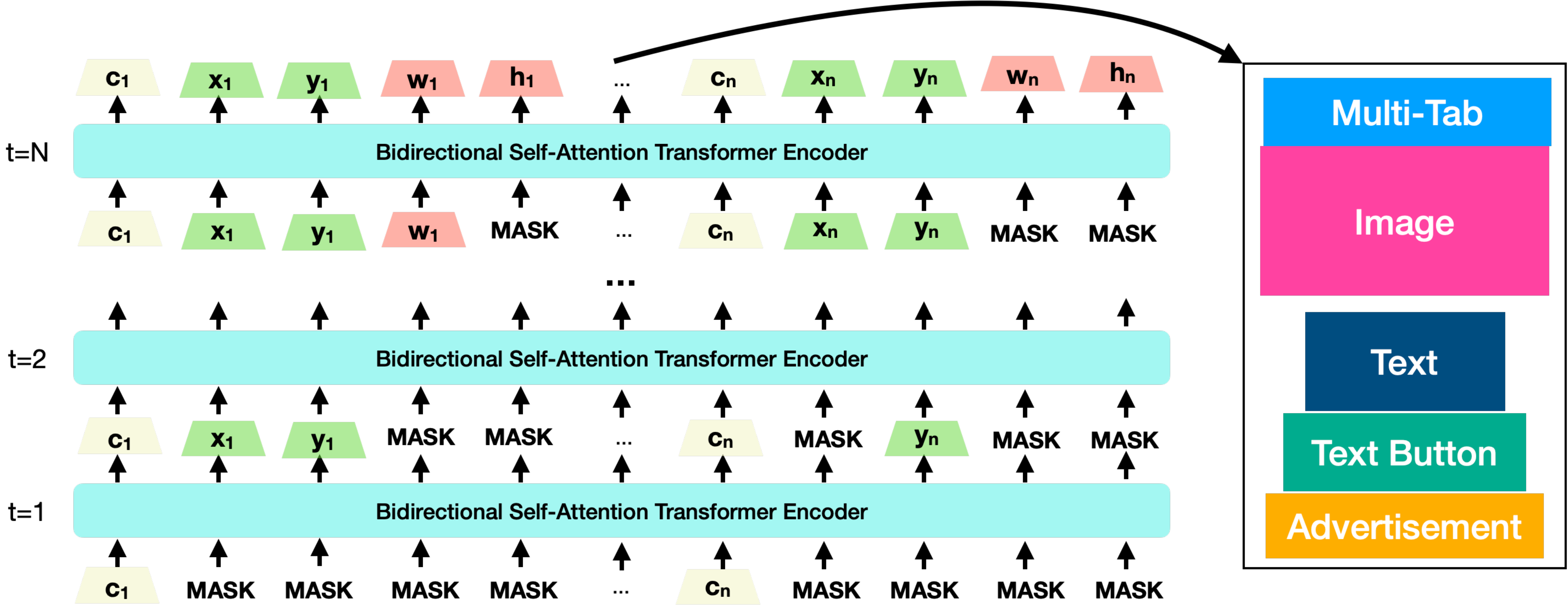}
   \caption{BLT Iterative Decoding Process.}
   \label{fig:blt_decode}
\end{subfigure}
\caption{The training (left) and decoding (right) stages of the proposed Bidirectional Layout Transformer (BLT).}
\label{fig:model}
\end{figure*}

The \modelname backbone is the multi-layer bidirectional Transformer encoder~\cite{vaswani2017attention} as shown in Fig.~\ref{fig:model}. We use the identical architecture as in the existing autoregressive layout transformers~\cite{arroyo2021variational,gupta2020layout} but a bidirectional attention mechanism.

Inspired by BERT~\cite{devlin2019bert}, during training, we randomly select a subset of attributes in the input sequence, replace them with a special ``$\textsc{[mask]}$'' token, and optimize the model to predict the masked attributes. For a layout sequence $l$, let $\mathcal{M}$ denote a set of masked positions. Replacing attributes in $l$ with ``$\textsc{[mask]}$'' at $\mathcal{M}$ yields the masked sequence $l^{\mathcal{M}}$.

Given a layout set $\mathcal{D}$, the training objective is to minimize the negative log-likelihood of the masked attributes:
\begin{equation}
\mathcal{L}_{mask} = - \mathop{\mathbb{E}}\limits_{l  \in \mathcal{D}} \Big[ \sum_{i\in \mathcal{M}} \log p(l_{i}|l^{\mathcal{M}}) \Big],
\end{equation}

The masking strategy greatly affects the quality of the masked language model~\cite{devlin2019bert}. BERT~\cite{devlin2019bert} applies random masking with a fixed ratio where a constant 15\% masks are randomly generated for each input. Similarly, we find masking strategy is important for layout generation, but the random masking used in BERT does not work well. We propose to use a new sampling policy. Specifically, we divide the attributes of an object into semantic groups, \eg Fig.~\ref{fig:model} showing 3 groups: category, position, and size. First, we randomly select a semantic group. Next, we dynamically sample the number of masked tokens from a uniform distribution between one and the number of attributes belonging to the chosen group, and then randomly mask that number of tokens in the selected group.  As such, it is guaranteed that the model only predicts attributes of the same semantic meaning each time. Therefore, given the hierarchical relations between these groups, we call this method as the hierarchical sampling. We will discuss how to apply the hierarchical sampling policy to decoding in the next subsection. 

\subsection{Parallel Decoding by Iterative Refinement}\label{sec:decoding}
In \modelname, all attributes in the layout are generated simultaneously in parallel. Since generating layouts in a single pass is challenging~\cite{ghazvininejad2019mask}, we employ a parallel language model. 
The core idea is to generate a layout iteratively in a small number of steps where parallel decoding is applied at each.

\begin{algorithm}[!ht]
  \caption{Decoding by Iterative Attribute Refinement}\label{alg:decoding}
  \begin{algorithmic}[1]
    \Require{Sequence $l$ with partially-known attributes. Constant $T$ for the number of iterations.}
    \For{$g$ in [$C$, $S$, $P$]} \Comment{Loop over semantic group}
    \For{$i \gets 1$ to $T/3$}                    
        \State {$p, l^{i}=\modelname~(l)$} 
        \State {$\gamma_{i} =\frac{T-3i}{T}$} \Comment{Compute mask ratio}
        \State {$n_{i} = \lfloor \gamma_{i} \times |g| \rfloor$} \Comment{$|g|$: \# attributes in $g$}
        \State {$\mathcal{M}= \arg_{k=n_i} \operatorname{top-k} (-p)$ \Comment{Get mask indices}
        }
        \State {Obtain $l$ by masking $l^{i}$ with respect to $\mathcal{M}$}
    \EndFor
    \EndFor
    \State \textbf{return} $l$
  \end{algorithmic}
\end{algorithm}

Algorithm~\ref{alg:decoding} presents the non-autoregressive decoding algorithm. The procedure is also illustrated in Fig.~\ref{fig:blt_decode}. The input to the decoding algorithm is a mixture sequence of known and unknown attributes, where the known attributes are given by the user inputs, and the model aims at generating the unknown attributes denoted by the $\textsc{[mask]}$ token.
Like in training, we employ the hierarchical sampling policy to generate attributes of three semantic groups: category ($C$), size ($S$), and position ($P$). For each iteration, one group of attributes is sampled. In Step 3 of Algorithm~\ref{alg:decoding}, the model makes parallel predictions for all unknown attributes, where $p$ denotes the prediction scores. Step 6 samples the attributes that belong to the selected group and have the lowest prediction scores. Finally, it masks low-confident attributes on which the model has doubts. The prediction
probabilities from the softmax layer are used as the confidence scores. These masked attributes will be re-predicted in the next iteration of decoding conditioning on all other ascertained attributes so far. The masking ratio calculated in Step 4 decreases with the number of iterations. This process will repeat $T$ times until all attributes of all objects are generated (\cf Fig.~\ref{fig:blt_decode}).

Our model is inspired by the autoregressive models in NLP~\cite{ghazvininejad2019mask,kong2020incorporating}. 
It is noteworthy that Algorithm~\ref{alg:decoding} differs from the non-autoregressive NLP models~\cite{ghazvininejad2019mask,kong2020incorporating} in the proposed hierarchical sampling. This paper finds applying~\cite{ghazvininejad2019mask,kong2020incorporating} to layout generation only leads to inferior results than our autoregressive baseline. We hypothesize that it is because layout attributes, unlike natural language, have apparent structures, and the non-autoregressive models designed for word sequences~\cite{ghazvininejad2019mask,kong2020incorporating} might not sufficiently capture the complex correlation between layout attributes.
We empirically demonstrate that Algorithm~\ref{alg:decoding} outperforms our non-autoregressive NLP baselines in Section~\ref{sec:order}.

Algorithm~\ref{alg:decoding} can be extended to unconditional generation. In this case, the input is a layout sequence of only ``$\textsc{[mask]}$'' tokens, and the same algorithm is used to generate all attributes in the layout. Unlike conditional generation, we need to know the sequence length in advance, \ie\ the number of objects to be generated. Here, we can use the prior distribution obtained on the training dataset. During decoding, we obtain the number of objects through sampling from this prior distribution.

\section{Experimental Results}\label{sec:exp}
This section verifies the proposed method on six diverse layout benchmarks under various metrics to examine realism, alignment, and semantic relevance. 
The results show our model performs favorably against the strong baselines and achieves a 4x-10x speedup than autoregressive decoding in layout generation.

\subsection{Setups}\label{sec:setups}
\paragraph{Datasets}
We employ six datasets that cover representative graphic design applications. \emph{RICO}~\cite{deka2017rico} is a dataset of user interface designs for mobile applications. It contains 91K entries with 27 object categories (button, toolbar, list item, \etc). 
\emph{PubLayNet}~\cite{zhong2019publaynet} contains 330K examples of machine annotated scientific documents crawled from the Internet. Its objects come from 5 categories: text, title, figure, list, and table. \emph{Magazine} \cite{zheng2019content} contains 4K images of magazine pages and six categories (texts, images, headlines, over-image texts, over-image headlines, backgrounds). \emph{Image Ads}~\cite{lee2020neural} is the commercial ads dataset with layout annotation detailed in~\cite{lee2020neural}.
\emph{COCO}~\cite{lin2014microsoft} contains  $\sim$100K images of natural scenes. We follow~\cite{arroyo2021variational} to use the Stuff variant, which contains 80 things and 91 stuff categories, after removing small bounding boxes ($\le$ 2\% image area), and instances tagged as ``iscrowd''. \emph{3D-FRONT}~\cite{fu20203dfront} is a repository of professionally designed indoor layouts. It contains around 7K room layouts with objects belonging to 37 categories, \eg, the table and bed. Different from previous datasets, objects in 3D-FRONT are represented by 3D bounding boxes. The maximum number of objects in our experiments is 25 in the RICO dataset and 22 in the PubLayNet dataset.

\paragraph{Evaluation metrics}
We employ five common metrics in the literature as well as a user study to validate the proposed method's effectiveness.
Specifically, \emph{IOU} measures the intersection over the union between the generated bounding boxes. We use an improved perceptual IOU (see more discussions in the Appendix). \emph{Overlap}~\cite{li2019layoutgan} measures the total overlapping area between any pair of bounding boxes inside the layout. \emph{Alignment}~\cite{lee2020neural} computes an alignment loss with the intuition that objects in graphic design are often aligned either by center or edge. \emph{FID}~\cite{heusel2017gans} measures the distributional distance of the generated layout to the real layout. Following~\cite{lee2020neural}, we compute FID using a binary layout classifier to discriminate real layouts. We employ a 2-layer Transformer to train the classifier.
Notice that the lower, the better for all IOU, Overlap, Alignment, and FID.

The above metrics ignore the input condition. For conditional generation, we employ a metric called \emph{Similarity}~\cite{patil2020read} and a user study, where the former compares the generated layout with the ground-truth layout under the same input. Following~\cite{patil2020read}, \emph{DocSim} is used to calculate the similarity between two layouts. The user study is used to further evaluate human's perception about the conditionally-generated layouts.

\paragraph{Generation settings} We examine three layout generation scenarios (2 conditional and 1 unconditional).
\begin{itemize}
 \setlength\itemsep{0em}
    \item Conditional on \textbf{Category}: only object categories are given by users. The model needs to predict the size and position of each object.
    \item Conditional on \textbf{Category + Size}: the object category and size are specified. The model needs to predict the positions, \ie placing objects on the canvas.
    \item \textbf{Unconditional} Generation: no information is provided by users. Prior layout transformer work focuses on this setting.
\end{itemize}
In unconditional generation, the model generates 1K samples from the random seed. The test split of each dataset is used for conditional generation.

\paragraph{Implementation details}
The model is trained for five trials with random initialization and the averaged metrics with standard deviations are reported. All models including ours have the same configuration, \ie, 4 layers, 8 attention heads, 512 embedding dimensions and 2,048 hidden dimensions. Adam optimizer~\cite{kingma2014adam} with $\beta_{1}=0.9$ and $\beta_{1}=0.98$ is used. Models are trained on 2$\times$2 TPU devices with batch size 64. 
For conditional generation, we randomly shuffle objects in the layout.
For unconditional generation, to improve diversity, we use the nucleus sampling~\cite{holtzman2019curious} with $p=0.9$ for the baseline Transformers and the top-k sampling ($k=5$) for our model. Greedy decoding method is used for conditional generation. Please refer to the Appendix for more detailed hyper-parameter configurations.

\subsection{Quantitative Comparison}
\begin{table*}[t]
    \begin{subtable}[h]{0.95\textwidth}
        \centering
        \scriptsize
        \begin{tabular*}{\textwidth}{@{\extracolsep{\fill}} lccccccc}
        \toprule
        RICO & \multicolumn{5}{c}{Conditioned on Category} & \multicolumn{2}{c}{+ Size}\\
        \cmidrule(lr){2-6} \cmidrule(lr){7-8}
        Model & IOU$\downarrow$ & Overlap$\downarrow$ & Alignment$\downarrow$ & FID$\downarrow$ & Sim.$\uparrow$ & Sim.$\uparrow$ &  FID$\downarrow$ \\
        \midrule
        L-VAE~\cite{jyothi2019layoutvae} & 0.41{\mytiny$\pm1.5\%$} & 0.39{\mytiny$\pm2.3\%$} & 0.38{\mytiny$\pm1.9\%$} & 122{\mytiny$\pm19$} & 0.13{\mytiny$\pm1.5\%$} & 0.19 & 76 \\
        NDN~\cite{lee2020neural}  & 0.37{\mytiny$\pm1.7\%$} & 0.36{\mytiny$\pm1.9\%$} & 0.41{\mytiny$\pm1.6\%$} & 97{\mytiny$\pm21$} & 0.15{\mytiny$\pm2.3\%$} & 0.21 & 63 \\
        Trans.~\cite{gupta2020layout} & 0.31{\mytiny$\pm0.2\%$}	& 0.33{\mytiny$\pm0.8\%$}	& 0.30{\mytiny$\pm0.8\%$}	 & 76{\mytiny$\pm24$} & 0.20{\mytiny$\pm0.1\%$} & -&-\\
        VTN~\cite{arroyo2021variational} &  \textbf{0.30}{\mytiny$\pm0.1\%$} & 0.30{\mytiny$\pm0.3\%$} & 0.32{\mytiny$\pm0.9\%$} & 82{\mytiny$\pm23$} & 0.20{\mytiny$\pm0.1\%$} & -&- \\
        Ours & \textbf{0.30}{\mytiny$\pm0.4\%$} & \textbf{0.23}{\mytiny$\pm0.2\%$} & \textbf{0.20}{\mytiny$\pm1.1\%$} & \textbf{70}{\mytiny$\pm29$} & \textbf{0.21}{\mytiny$\pm0.2\%$} & \textbf{0.30} & \textbf{26}\\
       \end{tabular*}
       \label{tab:condi_rico}
    \end{subtable}
    \hfill
    \begin{subtable}[h]{0.95\textwidth}
        \centering
        \scriptsize
        \begin{tabular*}{\textwidth}{@{\extracolsep{\fill}} l ccccccc}
        \toprule
        PubLayNet& \multicolumn{5}{c}{Conditioned on Category} & \multicolumn{2}{c}{+ Size}\\
        \cmidrule(lr){2-6} \cmidrule(lr){7-8}
        Model & IOU$\downarrow$ & Overlap$\downarrow$ & Alignment$\downarrow$ & FID$\downarrow$ & Sim.$\uparrow$ & Sim.$\uparrow$ &  FID$\downarrow$  \\
        \midrule
        L-VAE & 0.45{\mytiny$\pm1.3\%$} & 0.15{\mytiny$\pm0.9\%$} & 0.37{\mytiny$\pm0.7\%$} & 513{\mytiny$\pm26$} &  0.07{\mytiny$\pm0.3\%$} & 0.09 & 239 \\
        NDN & 0.34{\mytiny$\pm1.8\%$} & 0.12{\mytiny$\pm0.8\%$} & 0.39{\mytiny$\pm0.4\%$} & 425{\mytiny$\pm37$} &  0.06{\mytiny$\pm0.3\%$} & 0.09 & 178 \\
        Trans. &  \textbf{0.19}{\mytiny$\pm0.3\%$} & 0.06{\mytiny$\pm0.3\%$} & 0.33{\mytiny$\pm0.3\%$} & \textbf{127}{\mytiny$\pm29$} & \textbf{0.11}{\mytiny$\pm0.1\%$} & -&-\\
        VTN & 0.21{\mytiny$\pm0.6\%$} & 0.06{\mytiny$\pm0.2\%$} & 0.33{\mytiny$\pm0.4\%$} & 159{\mytiny$\pm21$} & 0.10{\mytiny$\pm0.1\%$} & -&-\\
        Ours & \textbf{0.19}{\mytiny$\pm0.2\%$} & \textbf{0.04}{\mytiny$\pm0.1\%$} & \textbf{0.25}{\mytiny$\pm0.7\%$} & 134{\mytiny$\pm24$}& \textbf{0.11}{\mytiny$\pm0.2\%$} & \textbf{0.18} & \textbf{87}\\
        \bottomrule
       \end{tabular*}
       \label{tab:condi_publaynet}
    \end{subtable}
     \caption{Conditional layout generation on two settings (Category and Category+ Size) on the large datasets of RICO and PubLayNet.}
     \label{tab:cond_rst}
\end{table*}

\begin{table*}[ht!]
    \begin{subtable}[h]{0.5\textwidth}
        \centering
        \scriptsize
        \begin{tabular*}{\textwidth}{@{\extracolsep{\fill}} lccc}
        \toprule
        COCO & \multicolumn{2}{c}{Conditioned on Category} & + Size\\
        \cmidrule(lr){2-3} \cmidrule(lr){4-4}
        Model & IOU$\downarrow$ & Sim.$\uparrow$ & Sim.$\uparrow$ \\
        \midrule
        Trans.~\cite{gupta2020layout}& 0.60{\mytiny$\pm0.4\%$}  & 0.20{\mytiny$\pm0.2\%$}  & -\\
        VTN~\cite{arroyo2021variational} & 0.63{\mytiny$\pm0.4\%$}  & 0.22{\mytiny$\pm0.1\%$}  & -\\
        Ours & \textbf{0.43}{\mytiny$\pm0.5\%$} & \textbf{0.24}{\mytiny$\pm0.1\%$}  & \textbf{0.44}\\
       \end{tabular*}
       \label{tab:condi_coco}
    \end{subtable}
    \hfill
    \begin{subtable}[h]{0.5\textwidth}
        \centering
        \scriptsize
        \begin{tabular*}{\textwidth}{@{\extracolsep{\fill}} lccc}
        \toprule
        Magazine & \multicolumn{2}{c}{Conditioned on Category} & + Size\\
        \cmidrule(lr){2-3} \cmidrule(lr){4-4}
        Model & IOU$\downarrow$ & Sim.$\uparrow$ & Sim.$\uparrow$ \\
        \midrule
        Trans. & 0.20{\mytiny$\pm0.8\%$} & 0.15{\mytiny$\pm0.3\%$}  & - \\
        VTN & \textbf{0.18}{\mytiny$\pm1.8\%$} & 	0.15{\mytiny$\pm0.9\%$}  & -\\
        Ours & \textbf{0.18}{\mytiny$\pm0.6\%$} & \textbf{0.18}{\mytiny$\pm0.4\%$} & \textbf{0.27} \\
       \end{tabular*}
       \label{tab:condi_magazine}
    \end{subtable}
    \hfill
    \begin{subtable}[h]{0.5\textwidth}
        \centering
        \scriptsize
        \begin{tabular*}{\textwidth}{@{\extracolsep{\fill}} lccc}
        \toprule
        Ads & \multicolumn{2}{c}{Conditioned on Category} & + Size\\
        \cmidrule(lr){2-3} \cmidrule(lr){4-4}
        Model & IOU$\downarrow$ & Sim.$\uparrow$ & Sim.$\uparrow$ \\
        \midrule
        Trans.~\cite{gupta2020layout} & 0.19{\mytiny$\pm0.1\%$} & 0.30{\mytiny$\pm0.1\%$}  & - \\
        VTN~\cite{arroyo2021variational} & 0.18{\mytiny$\pm0.2\%$}  & 0.30{\mytiny$\pm0.1\%$}  & - \\
        Ours & \textbf{0.10}{\mytiny$\pm0.4\%$}  & \textbf{0.31}{\mytiny$\pm0.1\%$}  & \textbf{0.41} \\
        \bottomrule
       \end{tabular*}
       \label{tab:condi_superbloom}
    \end{subtable}
    \hfill
    \begin{subtable}[h]{0.5\textwidth}
        \centering
        \scriptsize
        \begin{tabular*}{\textwidth}{@{\extracolsep{\fill}} lcc}
        \toprule
        3D-FRONT & Conditioned on Category & + size\\
        \cmidrule(lr){2-2} \cmidrule(lr){3-3}
        Model & Sim.$\uparrow$ & Sim.$\uparrow$ \\
        \midrule
        Trans. & 0.04{\mytiny$\pm0.7\%$}  & -\\
        VTN &  0.04{\mytiny$\pm0.4\%$}  & - \\
        Ours &  \textbf{0.06}{\mytiny$\pm0.7\%$}  & \textbf{0.10} \\
        \bottomrule
       \end{tabular*}
       \label{tab:front3d_rst}
    \end{subtable}
     \caption{Category (+ Size)  conditional layout generation on four datasets.}
     \label{tab:condi_rst_more}
\end{table*}

\paragraph{Conditional generation}
The results are shown in Table~\ref{tab:cond_rst} and Table~\ref{tab:condi_rst_more}. State-of-the-art layout transformers are compared \ie \textbf{LayoutTransformer (Trans.)}~\cite{gupta2020layout} and \textbf{Variational Transformer Network (VTN)}~\cite{arroyo2021variational}. In addition, two representative VAEs for conditional generation: \textbf{LayoutVAE (L-VAE)}~\cite{jyothi2019layoutvae} and \textbf{Neural Design Network (NDN)}~\cite{lee2020neural} are also compared on the large datasets of RICO and PubLayNet. Two conditional generation tasks are examined \ie Conditioned on Category and Conditioned on Category + Size (Column ``+ Size''). The same model is used for both conditional cases and ``-'' indicates the baseline models fail to process the condition ``Category + Size''. The results are aggregated on independently trained models, where the mean and standard deviation over five trails are reported.

Because of the non-autoregressive decoding, our model is able to conduct conditional generation on category + size while the baseline transformer models (Trans.~\cite{gupta2020layout} and VTN~\cite{arroyo2021variational}) fail. Our model also outperforms VAE-based conditional layout models (L-VAE~\cite{jyothi2019layoutvae} and NDN~\cite{lee2020neural}) across all metrics in Table~\ref{tab:cond_rst} by statistically significant margins. This result is consistent with the prior finding in~\cite{arroyo2021variational} that transformers outperform VAEs for unconditional layout generation.

\begin{table*}[t]
\centering
    \footnotesize
    \resizebox{\linewidth}{!}{
    \renewcommand{\tabcolsep}{8pt}
    \renewcommand{\arraystretch}{1.1}
    \begin{tabular}{@{}lcccccccccccccccc@{}}
        \toprule
        & \multicolumn{3}{c}{RICO} & \multicolumn{3}{c}{PubLayNet} & \multicolumn{3}{c}{COCO} \\
        \cmidrule(r){2-4}\cmidrule(l){5-7}\cmidrule(l){8-10} \cmidrule(l){11-13}
        Methods & IOU$\downarrow$ & Overlap$\downarrow$ & Alignment$\downarrow$ & IOU$\downarrow$ & Overlap$\downarrow$ & Alignment$\downarrow$ &  IOU$\downarrow$ & Overlap$\downarrow$ & Alignment$\downarrow$ \\
        \midrule
        LayoutVAE~\cite{jyothi2019layoutvae} & 0.193 & 0.400 & 0.416 & 0.171 & 0.321 & 0.472 & 0.325 & 2.819 & 0.246 \\
        Trans.~\cite{gupta2020layout} & 0.086 & 0.145 & 0.366& 0.039 & 0.006 & 0.361 & 0.194 & 1.709 & 0.334 \\
        VTN~\cite{arroyo2021variational} & 0.115 & 0.165 & 0.373 & 0.031 & 0.017 & 0.347 & 0.197 & 2.384 & 0.330\\
        Ours & 0.127 & 0.102 & 0.342 & 0.048 & 0.012 & 0.337 & 0.227 & 1.452 & 0.311 \\
        \bottomrule
    \end{tabular}
    }
\caption{Unconditional layout generation comparison to the state-of-the-art on three benchmarks. Results of baselines are cited from~\cite{arroyo2021variational} and our scores are calculated following the same method described in~\cite{arroyo2021variational}.}
\label{tab:uncond-all-merge}
\end{table*}

\paragraph{Unconditional Generation}
Although our model is not designed for this task, we compare it to the models~\cite{jyothi2019layoutvae,gupta2020layout,arroyo2021variational} on unconditional layout generation.
From Table~\ref{tab:uncond-all-merge}, our model outperforms LayoutVAE~\cite{jyothi2019layoutvae} and achieves comparable performance with two autoregressive transformers (Trans.~\cite{gupta2020layout} and VTN~\cite{arroyo2021variational}).

\begin{figure}[h]
	\centering
	\includegraphics[width=0.9\textwidth]{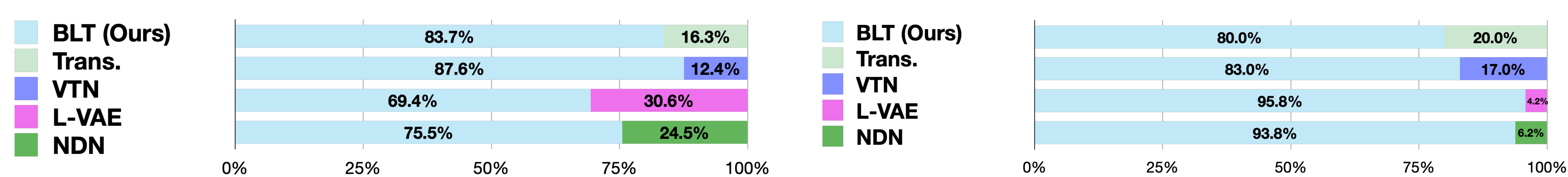}
	\caption{We conduct a user study to compare the quality of generated samples from our model and baseline models on RICO (left) and PubLayNet (Right).}
	\label{fig:user_study}
\end{figure}

\paragraph{User study} We conduct user studies on RICO and PubLayNet to assess generated layouts for conditional generation. We randomly select 50 generated layouts under both conditional settings specified in Section~\ref{sec:setups} and collect their golden layouts. For each trial, we present Amazon Mechanical Turk workers two layouts generated by different methods along with the golden layout for reference, and ask ``which layout is more similar to the true reference layout?''. There are 75 unique workers participating in the study. Qualitative comparison is shown in the Appendix. The results, which are plotted in Fig.~\ref{fig:user_study}, verify that the proposed model outperforms all baseline models for conditional layout generation.

\begin{figure*}[ht]
	\centering
	\includegraphics[width=\textwidth]{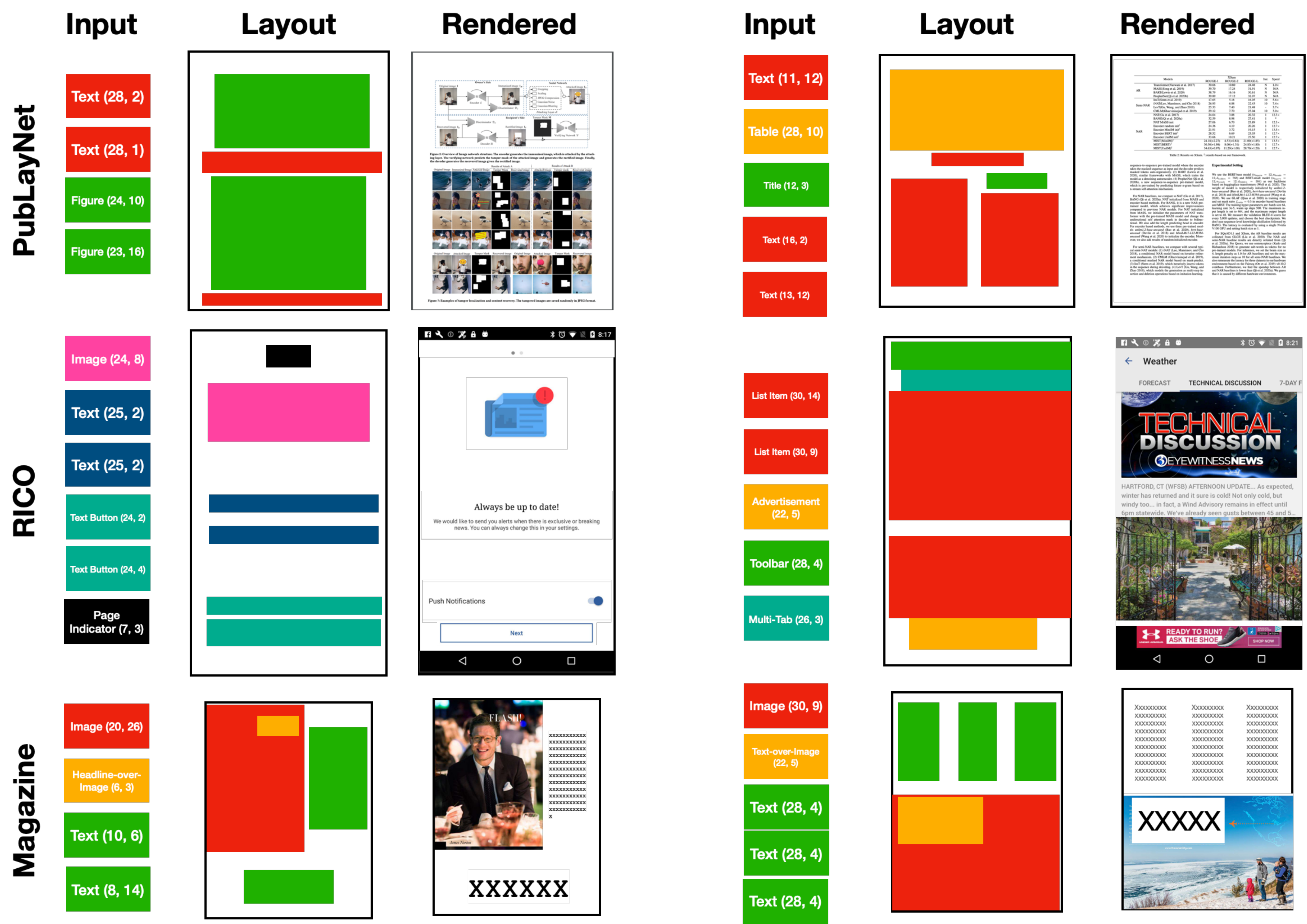}
	\caption{Conditional layout generation for scientific papers, user interface, and magazine. The user inputs are the object category and their size (width, height). We present the rendered examples constructed based on the generated layouts.}
	\label{fig:sample}
\end{figure*}

\begin{figure}[ht]
	\centering
	\includegraphics[width=0.7\textwidth]{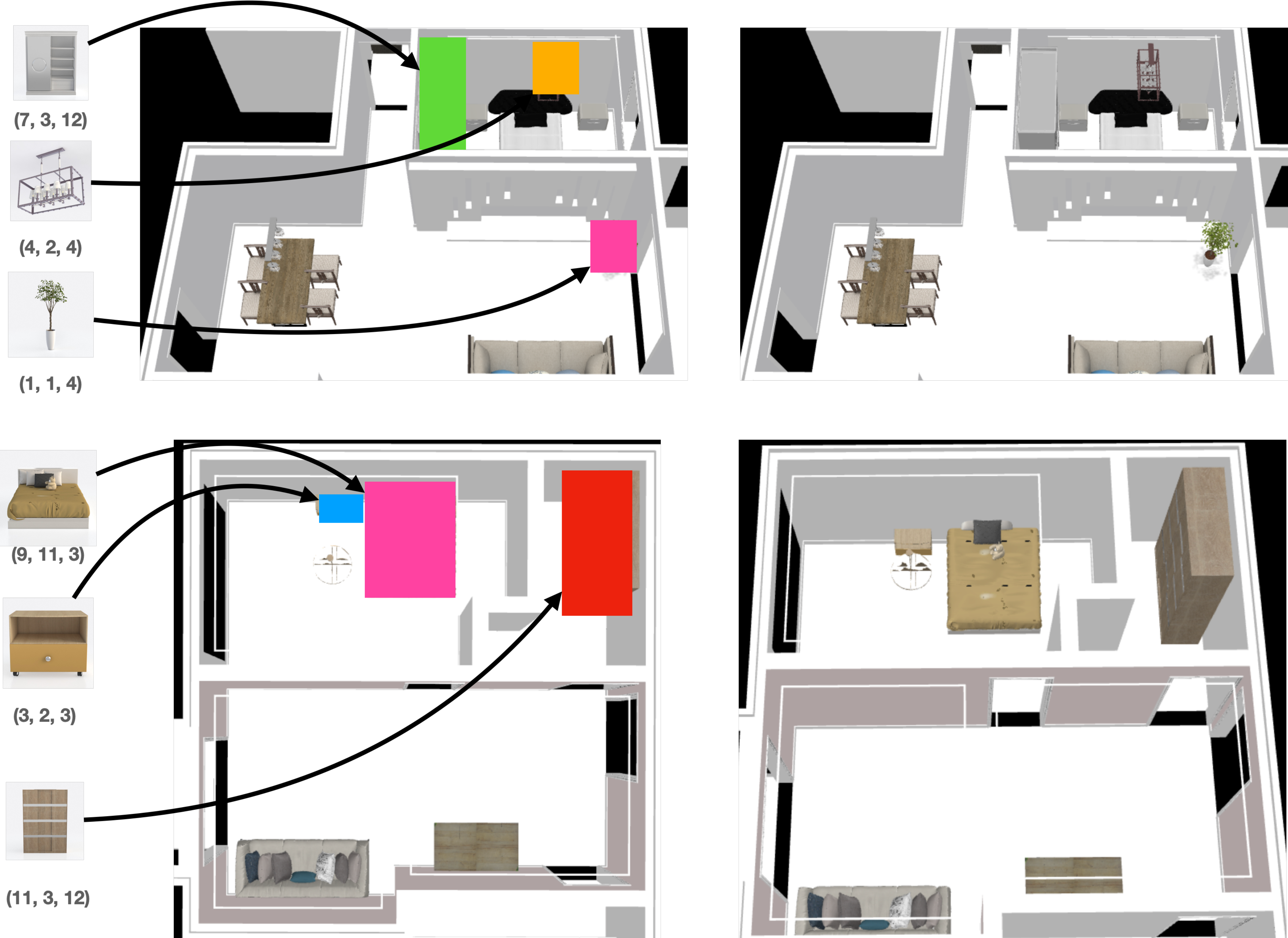}
	\caption{3D-FRONT sample layouts.}
	\label{fig:sample_3dfront}
\end{figure}

\begin{figure*}[t!]
    \centering
    \begin{subfigure}[t]{0.23\textwidth}
        \centering
        \includegraphics[width=0.95\textwidth]{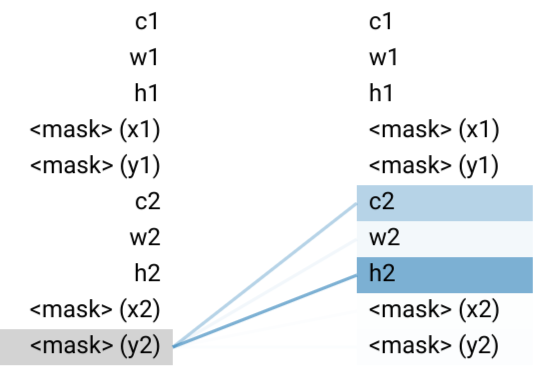}
        \caption{head 0-2}
    \end{subfigure}%
    ~ 
    \begin{subfigure}[t]{0.23\textwidth}
        \centering
        \includegraphics[width=0.95\textwidth]{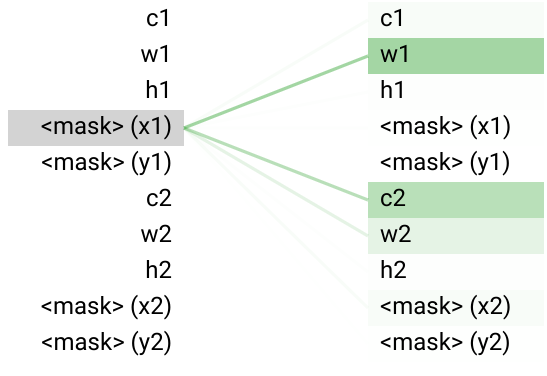}
        \caption{head 1-3}
    \end{subfigure}
    ~
    \begin{subfigure}[t]{0.23\textwidth}
        \centering
        \includegraphics[width=0.95\textwidth]{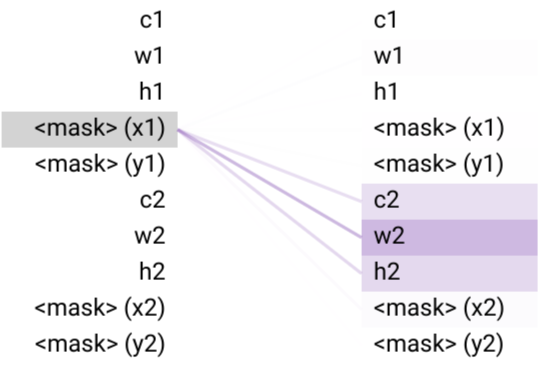}
        \caption{head 2-4}
    \end{subfigure}
    ~
    \begin{subfigure}[t]{0.23\textwidth}
        \centering
        \includegraphics[width=0.95\textwidth]{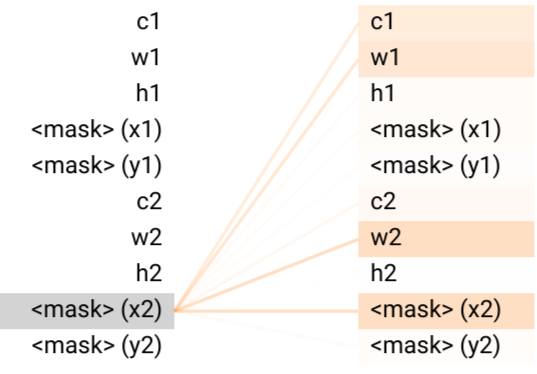}
        \caption{head 3-2}
    \end{subfigure}
    \caption{Examples of attention heads exhibiting the patterns for masked tokens. The darkness of a line indicates the strength of the attention weight (some attention weights are so low they are invisible). We use $\langle$layer$\rangle$-$\langle$head number$\rangle$ to denote a particular attention head.}
    \label{fig:head-vis}
\end{figure*}

\subsection{Qualitative Result}
We show some generated layouts, along with the rendered examples for visualization, in Fig.~\ref{fig:sample}. The setting is conditional generation on category and size for three design applications, including the mobile UI interface, scientific paper, and magazine. We observe that our method yields reasonable layouts, which facilitates generating high-quality outputs by rendering.

Next, we explore the home design task on the 3D-Front dataset~\cite{fu20203dfront}. The goal is to place the furniture with the user-given category and length, height, and width information. Examples are shown in Fig.~\ref{fig:sample_3dfront}. Unlike previous tasks, the model needs to predict the position of the 3D bounding box. The result suggests the feasibility of our method extending to 3D object attributes. The low similarity score on this dataset indicates that housing design layout is still a challenging task that needs future research.

To further understand what relationships between attributes \modelname has learned, we visualize the patterns in how our model's attention heads behave. We choose a simple layout with two objects and mask their positions ($x$, $y$). The model needs to predict these masked attributes from other known attributes. Examples of heads exhibiting these patterns are shown in Fig.~\ref{fig:head-vis}. We use $\langle$layer$\rangle$-$\langle$head number$\rangle$ to denote a particular attention head. For the head 0-2, $\textsc{[mask]}_{y_{2}}$ specializes to attending on its category (c2) and especially, its height information (h2), which is reasonable because $y$-coordinate is highly relevant to the height of the object. Furthermore, for heads 2-4 and 3-2, $\textsc{[mask]}_{x_{1}}$ focuses on the width of not only the first but the second object as well. Given this contextual information from other objects, the model is able to predict the position of these objects more accurately. The similar pattern is also found at head 3-2 for $\textsc{[mask]}_{x_{2}}$.

\subsection{Ablation Study}\label{sec:order}

\paragraph{Decoding speed}\label{sec:speed}
\begin{figure}[!ht]
\begin{minipage}[c]{0.3\textwidth}
    \centering
	\includegraphics[width=0.95\textwidth]{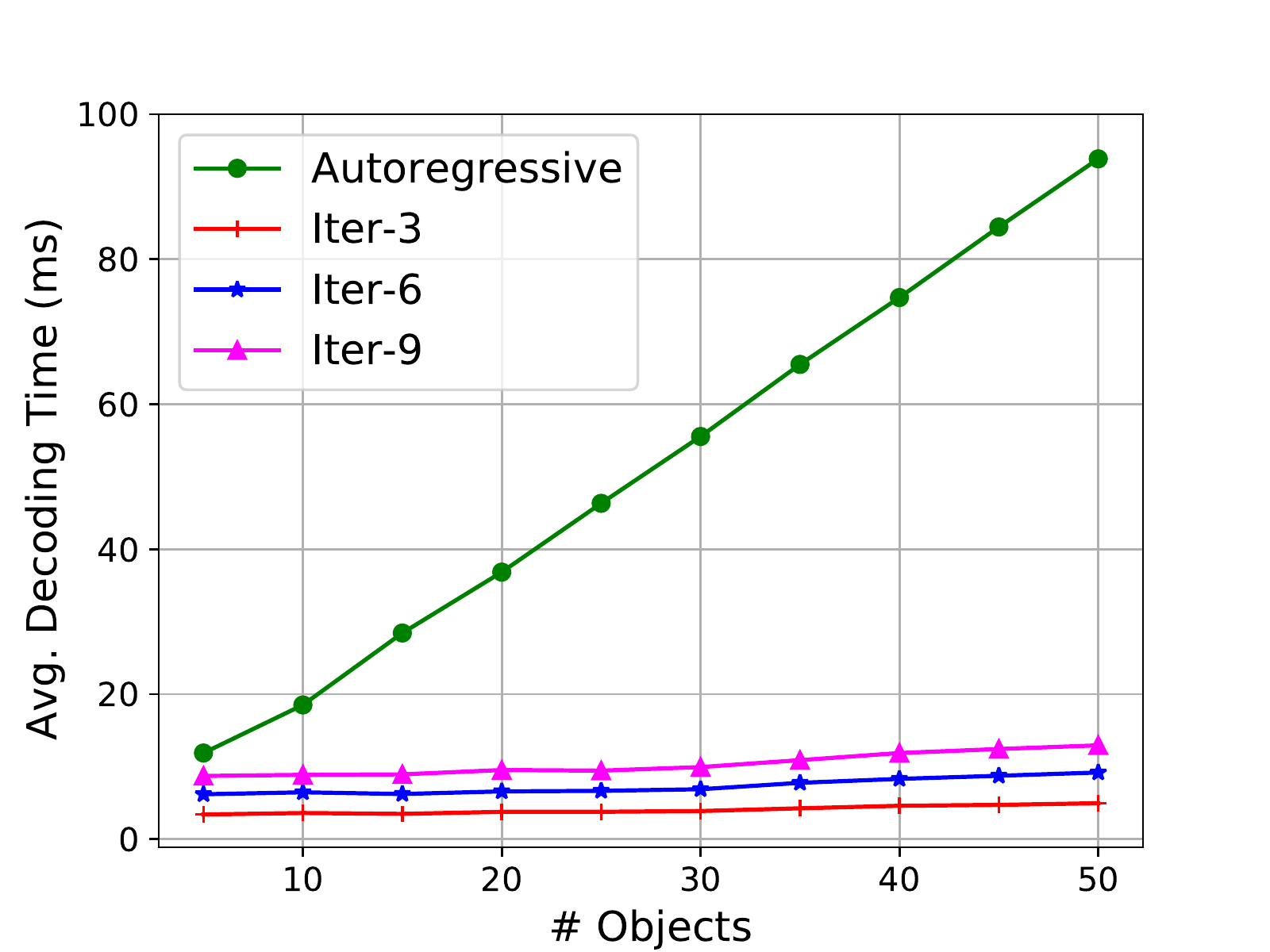}
  \end{minipage}\hfill
  \begin{minipage}[c]{0.6\textwidth}
    \caption{Decoding speed versus number of generated assets.  `Autoregressive' denote the autogressive Transformer-based model~\cite{gupta2020layout}. `Iter-*` shows the proposed model with various number of iterations.}\label{fig:speed}
  \end{minipage} 
\end{figure}
We compare the inference speed of our model and the autoregressive transformer models~\cite{gupta2020layout,arroyo2021variational}. Specifically, all models generate 1,000 layouts with batch size 1 on a single GPU. The average decoding time in millisecond is reported. The result is shown in Fig.\ref{fig:speed}, where the $x$-axis denotes the number of objects in the layout. 
It shows that autoregressive decoding time grows with \#objects. On the contrary, the decoding speed of the proposed model appears not affected by \#objects. The speed advantage becomes evident when producing dense layouts. For example, our fastest model obtains a 4x speedup when generating around 10 objects and a 10x speedup for 20 objects. 

\paragraph{Hierarchical sampling} This experiment investigates the effectiveness of the hierarchical sampling strategy used in training (Section~\ref{sec:training}) and non-autoregressive decoding (Section~\ref{sec:decoding}). Specifically, we compare with the non-autoregressive method~\cite{ghazvininejad2019mask} in NLP on the large datasets of RICO and PubLayNet in Table~\ref{tab:sampling}. Autoregressive transformer results~\cite{gupta2020layout} are also included for reference but notice that autoregressive methods~\cite{gupta2020layout} have difficulties with conditional generation.

The results in Table~\ref{tab:sampling} show that the non-autoregressive baseline yields inferior results than the autoregressive one. We hypothesize that it is because the non-autoregressive models designed for word sequences~\cite{ghazvininejad2019mask,kong2020incorporating} might not sufficiently capture the apparently-structural correlation between layout attributes. The proposed method with hierarchical sampling significantly outperforms the non-autoregressive NLP baseline, which suggests the necessity of the proposed hierarchical sampling strategy. We also explore the effect of hierarchical sampling order. In Algorithm~\ref{alg:decoding}, we prespecify an order of attribute groups, \ie, Category (C) $\rightarrow$ Size (S) $\rightarrow$ Position (P). Here, more orders are explored in Table~\ref{tab:attr-order}.
It seems better to first generate the category and afterward determine either location or size.
\begin{table}[t]
\centering
\footnotesize
\begin{tabular}{cccccc}
\toprule
RICO &IoU $\downarrow$ & Overlap$\downarrow$ & Align.$\downarrow$ & FID$\downarrow$ & Sim. $\uparrow$\\
\midrule
Autoregressive~\cite{gupta2020layout}  & \textbf{0.30} & 0.33 & 0.30 &  76 & 0.20 \\
\midrule
Non-autoregressive~\cite{ghazvininejad2019mask} & 0.37 & 0.33 & 0.24 &  104 & 0.17\\
Non-autoregressive + HSP (Ours) & \textbf{0.30} & \textbf{0.23} & \textbf{0.20} & \textbf{70} & \textbf{0.21} \\
\toprule
PubLayNet &IoU $\downarrow$ & Overlap$\downarrow$ & Align.$\downarrow$ & FID$\downarrow$ & Sim. $\uparrow$ \\
\midrule
Autoregressive~\cite{gupta2020layout} & \textbf{0.19} & 0.06 & 0.33 & \textbf{127} & \textbf{0.11} \\
\midrule
Non-autoregressive~\cite{ghazvininejad2019mask} & 0.16 & 0.12 & 0.32 & 217 & 0.09 \\
Non-autoregressive + HSP (Ours) & \textbf{0.19} & \textbf{0.04} & \textbf{0.25} & 134 & \textbf{0.11}\\
\bottomrule
\end{tabular}
\caption{Comparison with the non-autoregressive method~\cite{ghazvininejad2019mask} in NLP on the RICO and PubLayNet datasets. Autoregressive results are included for reference. HSP denotes hierarchical sampling policy proposed in this work.
}
\label{tab:sampling}
\end{table}

\begin{table}[t]
\begin{minipage}[c]{0.3\textwidth}
\centering
\scriptsize
\begin{tabular}{cccc}
\toprule
Order &IoU $\downarrow$ & Overlap$\downarrow$ & Alignment$\downarrow$ \\
\midrule
C$\rightarrow$S$\rightarrow$P & \textbf{0.127} & \textbf{0.102} & \textbf{0.342}\\
C$\rightarrow$P$\rightarrow$S & 0.129 & 0.107 & 0.344 \\
S$\rightarrow$C$\rightarrow$P & 0.147 & 0.109 & 0.351 \\
S$\rightarrow$P$\rightarrow$C & 0.162 & 0.121 & 0.357 \\
\bottomrule
\end{tabular}
\end{minipage}\hfill
\begin{minipage}[c]{0.6\textwidth}
\caption{Layout generation results with different iteration group orders on the RICO dataset. C, S, and P denote category, size, and position attribute groups, respectively.
}
\label{tab:attr-order}
\end{minipage}
\end{table}

\section{Conclusion and Future Work}
We present \modelname, a bidirectional layout transformer capable of empowering the transformer-based models to carry out conditional and controllable layout generation. Moreover, we propose a hierarchical sampling policy during BLT training and inference processes which has been shown to be essential for producing high-quality layouts. Thanks to the high computation parallelism, BLT achieves 4-10 times speedup compared to the autoregressive transformer baselines during inference. Experiments on six benchmarks show the effectiveness and flexibility of \modelname. A limitation of our work is content-agnostic generation. We leave this out to have a fair and lateral comparison to our baselines which do not use visual information either. In the future, we will explore using rich visual information.
\section*{Acknowledgement}
The authors would like to thank all anonymous reviewers and area chairs for helpful comments.

%
%
\bibliographystyle{splncs04}
\bibliography{main}

\newpage
\section*{Appendix}
\appendix

\section{Implementation Details}
\paragraph{Training.} To find out the optimal hyperparameters for each task, we use a grid search for the following ranges
of possible values, learning rate in $\{1e-3, 3e-3, 5e-3\}$, dropout and attention dropout in $\{0.1, 0.3\}$.
The data preprocessing procedure discussed in~\cite{arroyo2021variational} is used. 
All baseline models, including ours, are trained on the same dataset by five independent trials, where the averaged metrics with standard deviations are reported.  

\begin{figure}[h!]
	\centering
	\includegraphics[height=0.2\textwidth, width=0.3\textwidth]{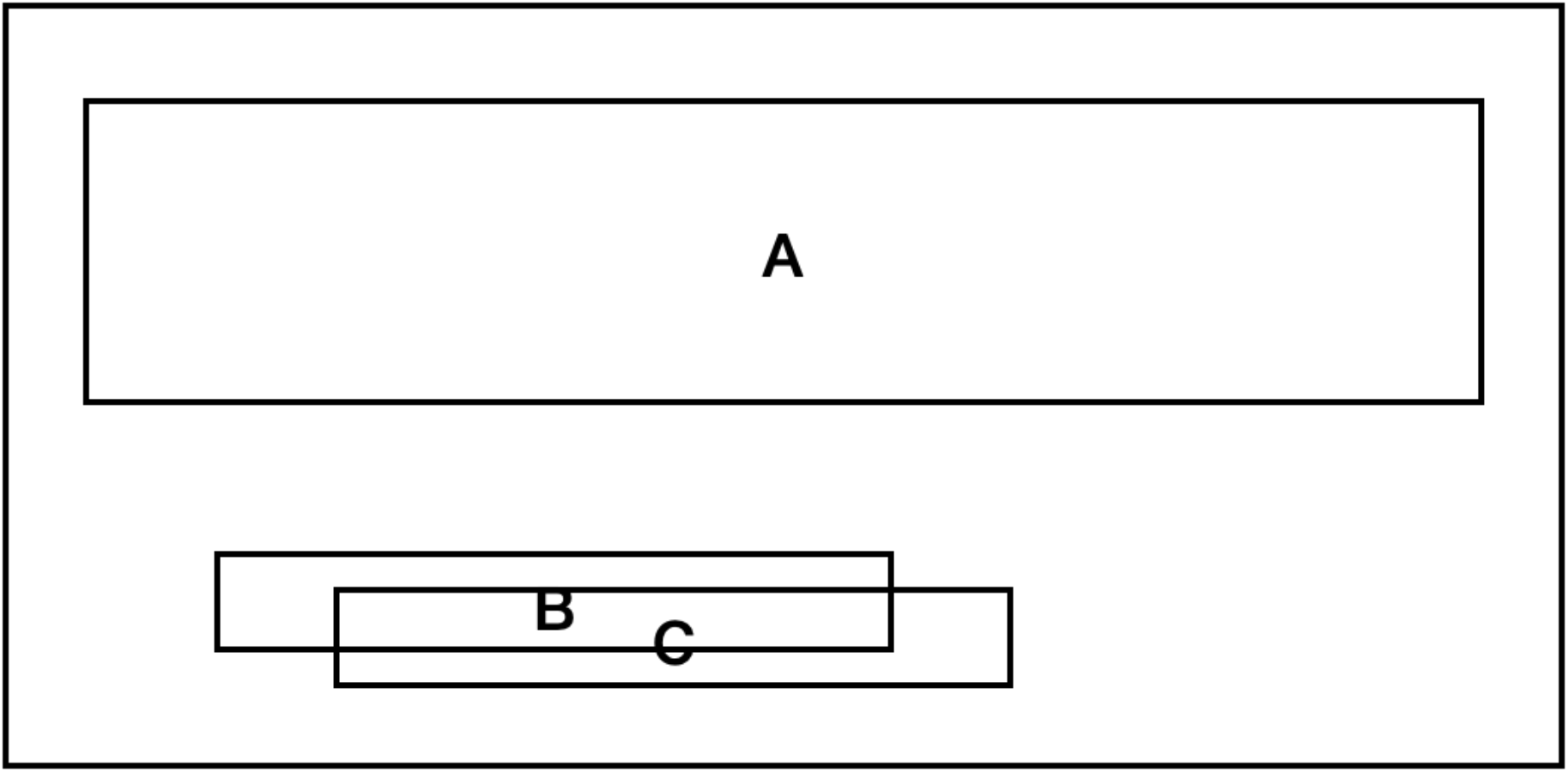}
	\caption{A toy layout sample for the IOU computation. The metrics used yields more reasonable IOU $\frac{0.5}{6.5}=\frac{1}{13}$ than the IOU $\frac{0.5}{1.5}=\frac{1}{3}$ used in~\cite{arroyo2021variational} }
	\label{fig:iou}
\end{figure}

\paragraph{Notes on the evaluation metric IOU:} In~\cite{arroyo2021variational}, the author calculates the IOU scores between all pairs of overlapped objects and average them. In our work, we propose to use the so-called perceptual IOU score which first projects the layouts as if they were images then computes the ovarlapped area divided by \textbf{the union area of all objects}.
We show the difference via a toy example in Fig.~\ref{fig:iou}. The areas of objects $A$, $B$ and $C$ are 5, 1, 1, the overlapped area of $B$ and $C$ are 0.5.
Based on the IOU computation in~\cite{arroyo2021variational}, since they just care about overlapped objects, only the IOU of objects \textbf{B} and \textbf{C} are computed which is $\frac{0.5}{1.5}=\frac{1}{3}$. On the contrary, in our IOU computation, the overlapped area of $B$ and $C$ will be divided the union area of all objects, hence, the IOU of this layout is $\frac{0.5}{6.5}=\frac{1}{13}$ which is more reasonable than their result.

\section{Additional Quantitative Results}
We show more results with more metrics on CoCo, Magazine and Ads datasets. Our proposed model consistently achieve better results on these datasets compared to autoregressive transformer-based models, demonstrating the effectiveness of our model.

\begin{table*}[ht!]
\centering
    \begin{subtable}[h]{0.95\textwidth}
        \centering
        \scriptsize
        \begin{tabular*}{\textwidth}{@{\extracolsep{\fill}} lccccc}
        \toprule
        COCO & \multicolumn{4}{c}{Conditioned on Category} & + Size\\
        \cmidrule(lr){2-5} \cmidrule(lr){6-6}
        Model & IOU$\downarrow$ & Overlap$\downarrow$ & Alignment$\downarrow$ & Sim.$\uparrow$ & Sim.$\uparrow$ \\
        \midrule
        Trans.& 0.60{\mytiny$\pm0.4\%$}  & \textbf{1.66}{\mytiny$\pm2.0\%$}  & 0.34{\mytiny$\pm0.2\%$}  & 0.20{\mytiny$\pm0.2\%$}  & -\\
        VTN & 0.63{\mytiny$\pm0.4\%$}  & 1.79{\mytiny$\pm1.6\%$}  & 0.32{\mytiny$\pm0.3\%$}  & 0.22{\mytiny$\pm0.1\%$}  & -\\
        Ours & \textbf{0.35}{\mytiny$\pm0.5\%$}  & 1.93{\mytiny$\pm5.0\%$}  & \textbf{0.16}{\mytiny$\pm0.5\%$}  & \textbf{0.24}{\mytiny$\pm0.1\%$}  & \textbf{0.44}\\
        \midrule
      \end{tabular*}
    \end{subtable}
    \begin{subtable}[h]{0.95\textwidth}
        \centering
        \scriptsize
        \begin{tabular*}{\textwidth}{@{\extracolsep{\fill}} lccccc}
        \toprule
        Magazine& \multicolumn{4}{c}{Conditioned on Category} & + Size\\
        \cmidrule(lr){2-5} \cmidrule(lr){6-6}
        Model & IOU$\downarrow$ & Overlap$\downarrow$ & Alignment$\downarrow$ & Sim.$\uparrow$ & Sim.$\uparrow$ \\
        \midrule
        Trans. & 0.20{\mytiny$\pm0.8\%$}  & 0.22{\mytiny$\pm1.6\%$}  & 0.48{\mytiny$\pm1.1\%$}  & 0.15{\mytiny$\pm0.3\%$}  & - \\
        VTN & \textbf{0.18}{\mytiny$\pm1.8\%$}  &	0.15{\mytiny$\pm1.2\%$}  & 0.47{\mytiny$\pm1.4\%$}  & 	0.15{\mytiny$\pm0.9\%$}  & -\\
        Ours & \textbf{0.18}{\mytiny$\pm0.6\%$}  & \textbf{0.12}{\mytiny$\pm1.8\%$}  & \textbf{0.44}{\mytiny$\pm1.9\%$} & \textbf{0.18}{\mytiny$\pm0.4\%$} & \textbf{0.27} \\
        \midrule
      \end{tabular*}
    \end{subtable}
    \begin{subtable}[h]{0.95\textwidth}
        \centering
        \scriptsize
        \begin{tabular*}{\textwidth}{@{\extracolsep{\fill}} lccccc}
        \toprule
        Ads & \multicolumn{4}{c}{Conditioned on Category} & + Size\\
        \cmidrule(lr){2-5} \cmidrule(lr){6-6}
        Model & IOU$\downarrow$ & Overlap$\downarrow$ & Alignment$\downarrow$ & Sim.$\uparrow$ & Sim.$\uparrow$ \\
        \midrule
        Trans & 0.19{\mytiny$\pm0.1\%$}  & 0.15{\mytiny$\pm0.1\%$}  & 0.35{\mytiny$\pm0.1\%$}  & 0.30{\mytiny$\pm0.1\%$}  & - \\
        VTN & 0.18{\mytiny$\pm0.2\%$}  & 0.15{\mytiny$\pm0.1\%$}  & 0.33{\mytiny$\pm0.1\%$}  & 0.30{\mytiny$\pm0.1\%$}  & - \\
        Ours & \textbf{0.10}{\mytiny$\pm0.4\%$}  & \textbf{0.10}{\mytiny$\pm0.4\%$}  & \textbf{0.18}{\mytiny$\pm0.6\%$}  & \textbf{0.31}{\mytiny$\pm0.1\%$}  & 0.41 \\
        \bottomrule
      \end{tabular*}
    \end{subtable}
    
     \caption{Category (+ Size)  conditional  layout generation performance on various benchmarks.}
     \label{tab:cond_rst_supp}
\end{table*}

\begin{figure}[ht!]
    \centering
    \begin{subfigure}[t]{0.4\textwidth}
        \centering
        \includegraphics[width=0.98\textwidth]{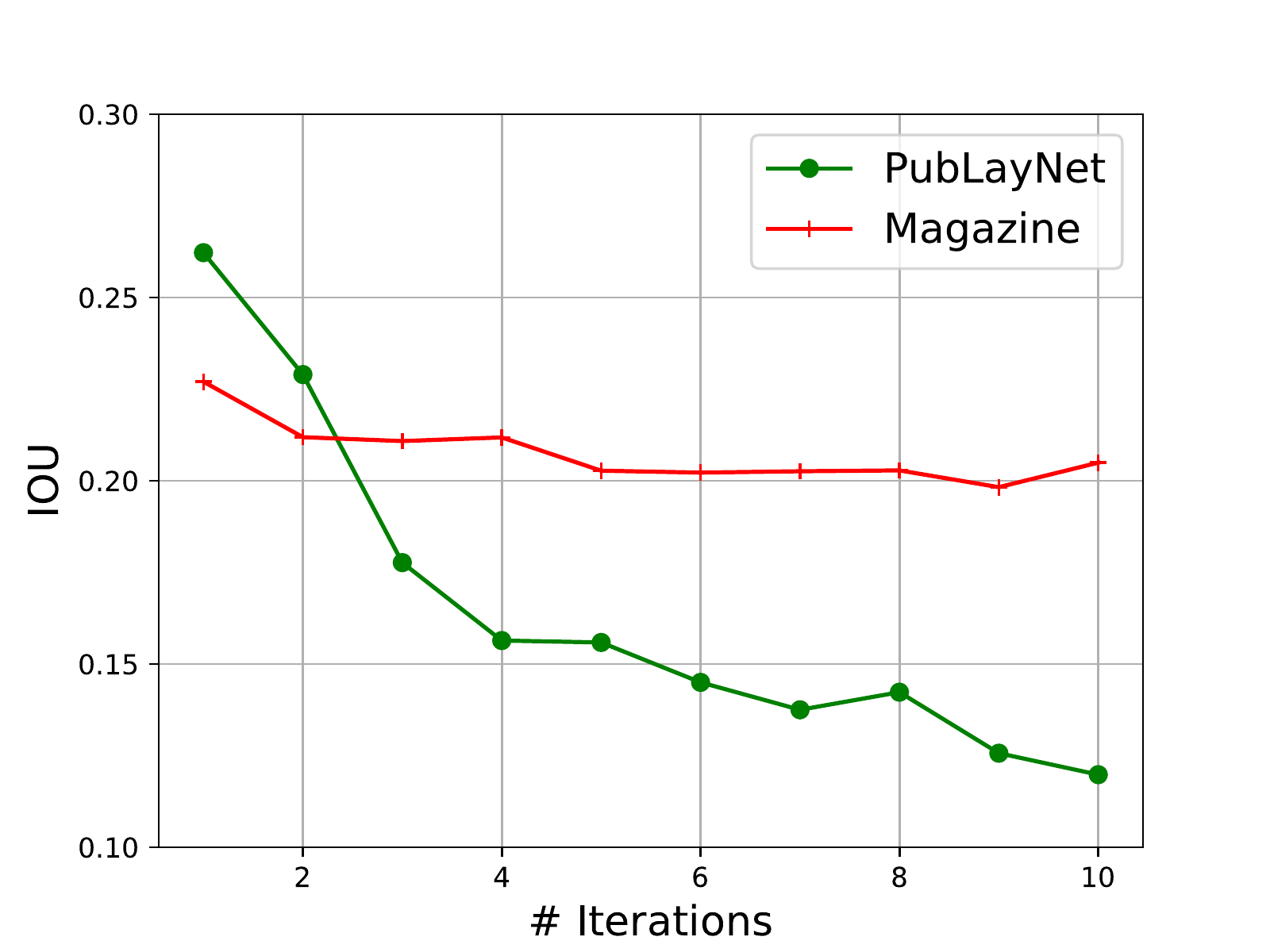}
    \end{subfigure}%
    ~ 
    \begin{subfigure}[t]{0.4\textwidth}
        \centering
        \includegraphics[width=0.98\textwidth]{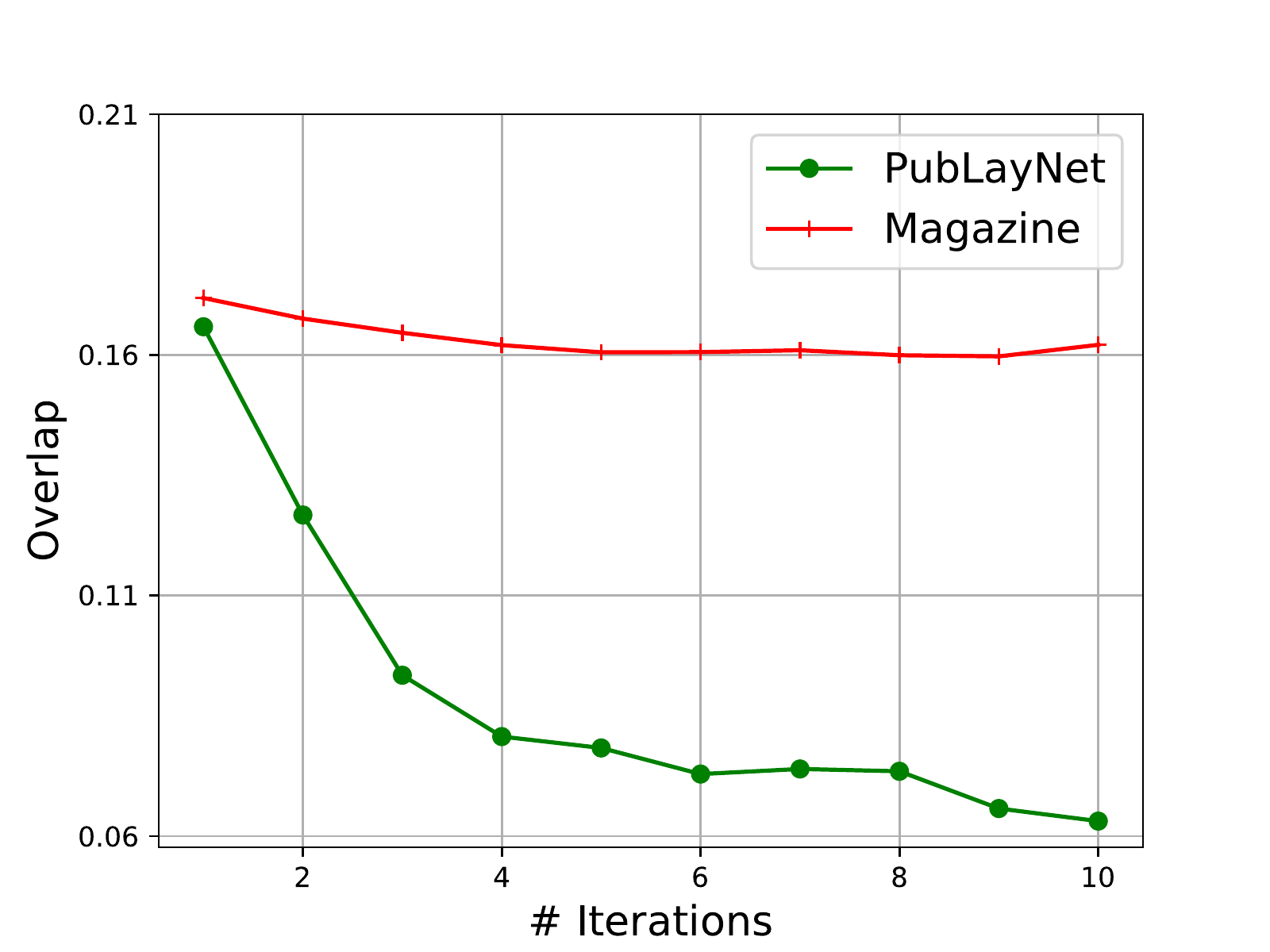}
    \end{subfigure}%
    \caption{IOU and overlap scores at different decoding iterations on two datasets.}
    \label{fig:refine_scores}
\end{figure}

\section{Additional Visualization}

\begin{figure*}[h!]
	\centering
	\includegraphics[width=0.95\textwidth]{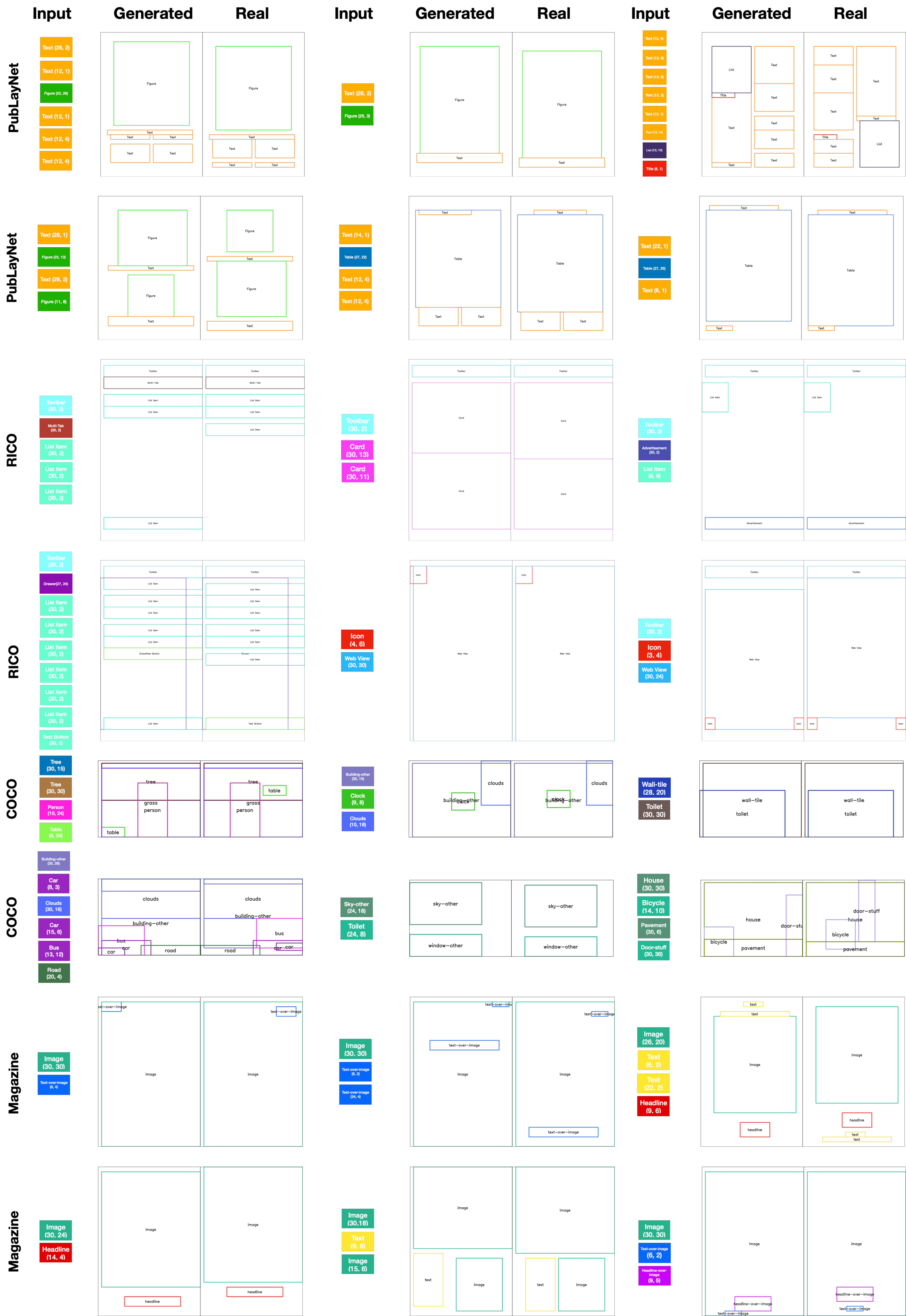}
	\caption{Conditional layout generation for scientific papers, user interface, and magazine. The user inputs are the object category and their size (width, height). We compare the generated layout and the real layout with the same input in the dataset.}
	\label{fig:more_sample}
\end{figure*}

\begin{figure*}[h!]
	\centering
	\includegraphics[width=0.95\textwidth]{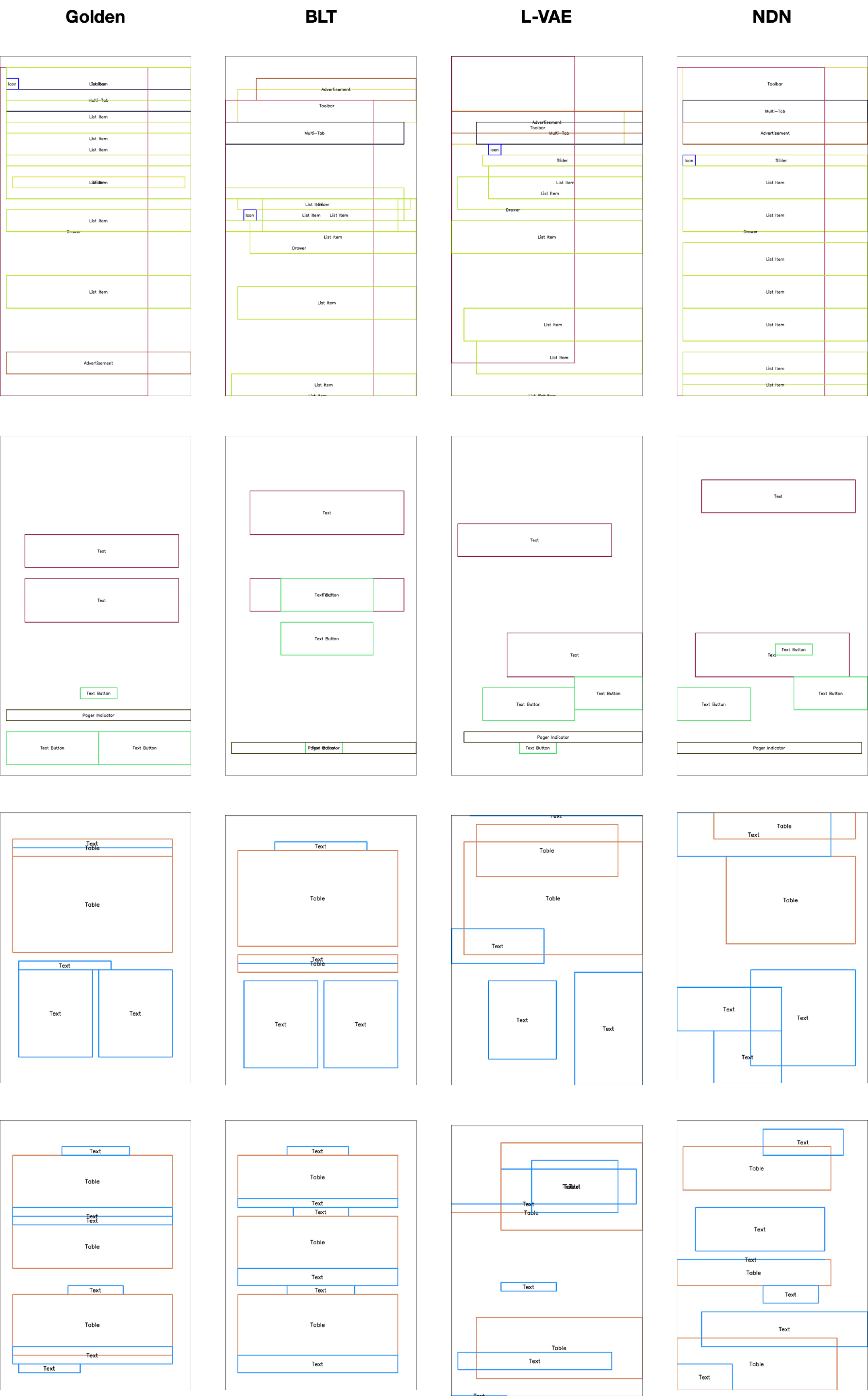}
	\caption{Qualitative Results for conditional generation on PublayNet and RICO from BLT and VAE-based models.}
	\label{fig:blt_vae}
\end{figure*}

\subsection{Qualitative Results for Conditional Generation}
We show samples in Fig.~\ref{fig:more_sample} from conditional generation on category and size for four design applications including the mobile UI interface, scientific paper, magazine and natural scenes. We also show some samples in comparison with Layout-VAE~\cite{jyothi2019layoutvae} and NDN~\cite{lee2020neural} in Fig.~\ref{fig:blt_vae}.

\begin{figure*}[!ht]
	\centering
	\includegraphics[width=0.8\textwidth]{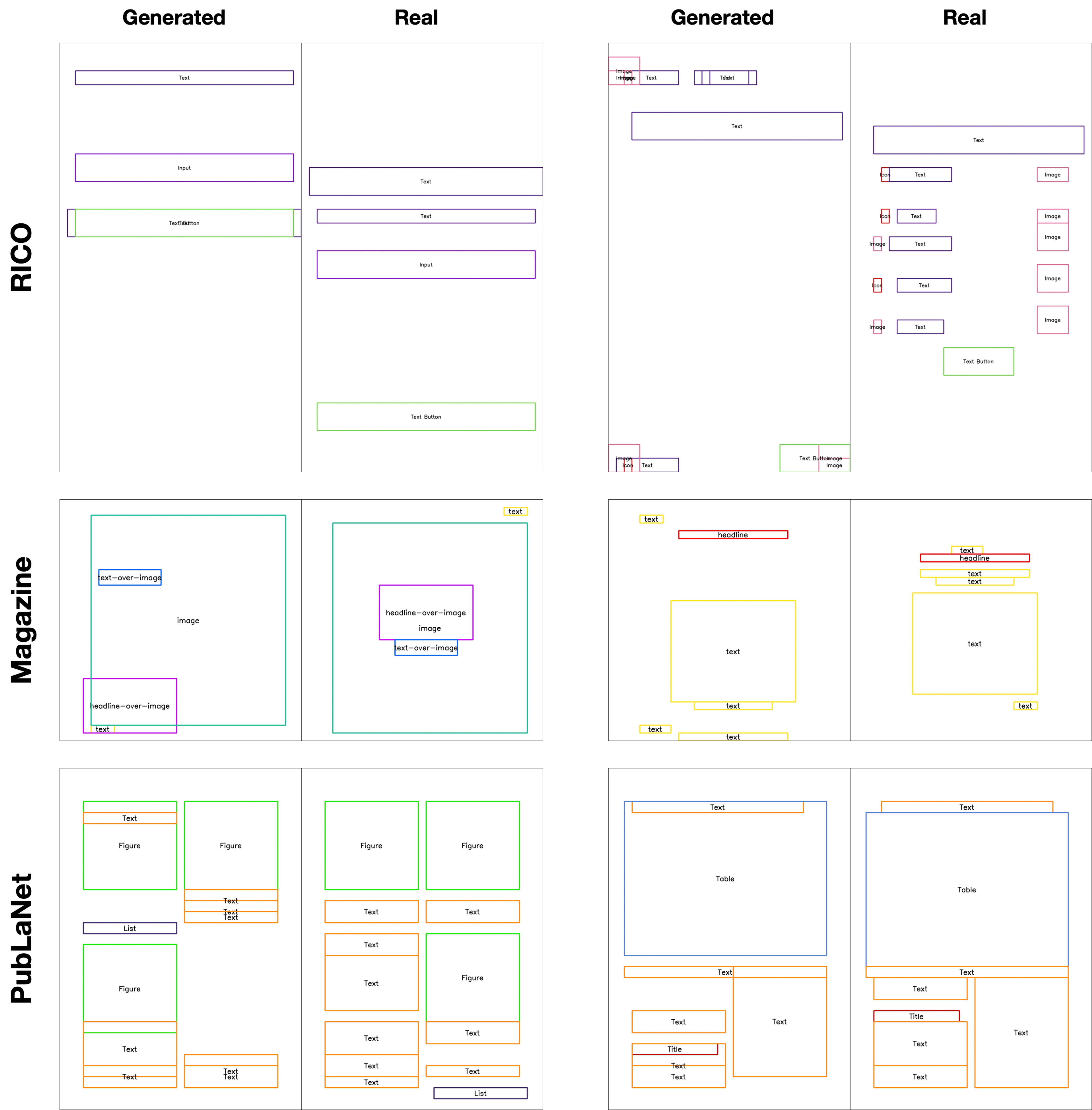}
	\caption{Failure cases for layout generation using the propose method. We compare the generated layout and the real layout with the same input in the dataset. See Section~\ref{sec:failure_cases} for more discussion.}
	\label{fig:failure}
\end{figure*}

\subsection{Examples of Diverse Conditional Generation}
\begin{figure*}[h!]
	\centering
	\includegraphics[width=0.95\textwidth]{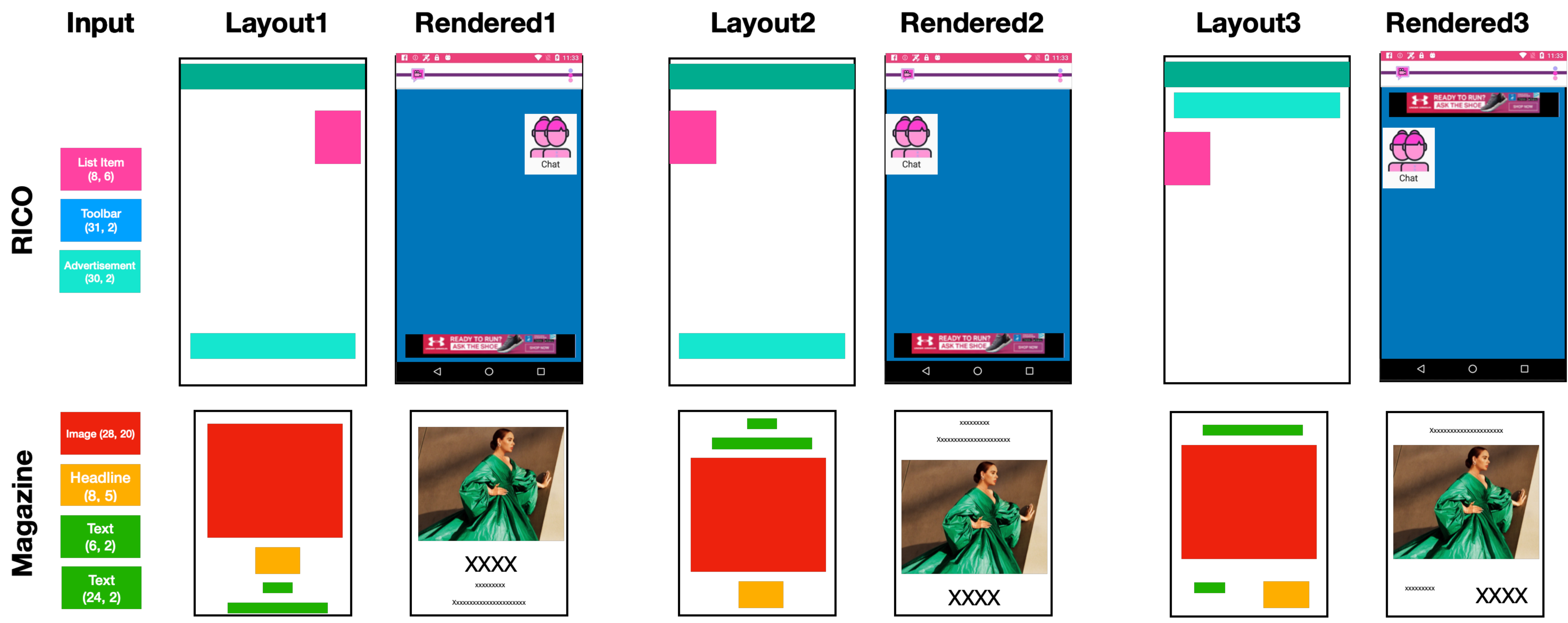}
	\caption{Diverse conditional generation via top-$k$ sampling method.}
	\label{fig:diverse_sample}
\end{figure*}

In our main experiment, we use greedy search to find out the most likely candidate for each attribute at each iteration. Here, we generate layouts through sampling the top-k ($k=10$) from the likelihood distribution for category + size conditional generation. This leads to diverse layouts. Some examples are shown in Fig.~\ref{fig:diverse_sample}.

\subsection{More Attention Head Patterns}
\begin{figure*}[h!]
    \centering
    \begin{subfigure}[t]{0.23\textwidth}
        \centering
        \includegraphics[width=0.95\textwidth]{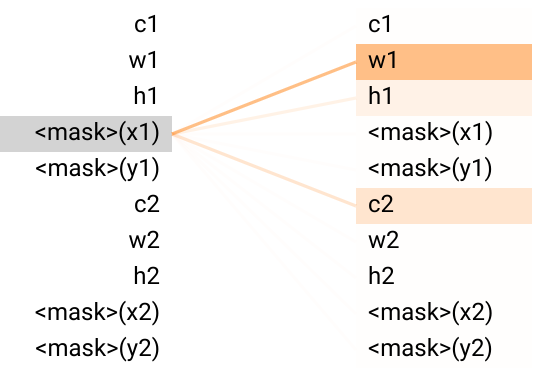}
        \caption{head 1-1}
    \end{subfigure}%
    ~ 
    \begin{subfigure}[t]{0.23\textwidth}
        \centering
        \includegraphics[width=0.95\textwidth]{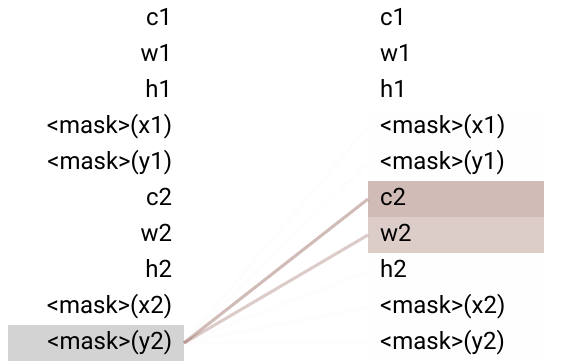}
        \caption{head 1-3}
    \end{subfigure}
    ~
    \begin{subfigure}[t]{0.23\textwidth}
        \centering
        \includegraphics[width=0.95\textwidth]{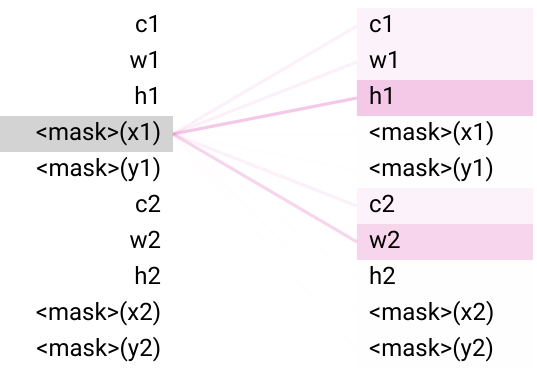}
        \caption{head 1-6}
    \end{subfigure}
    ~
    \begin{subfigure}[t]{0.23\textwidth}
        \centering
        \includegraphics[width=0.95\textwidth]{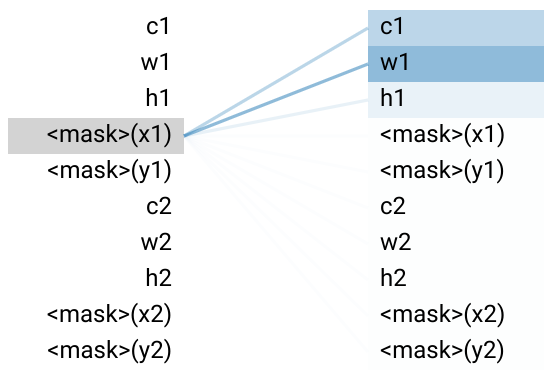}
        \caption{head 0-4}
    \end{subfigure}
    \begin{subfigure}[t]{0.23\textwidth}
        \centering
        \includegraphics[width=0.95\textwidth]{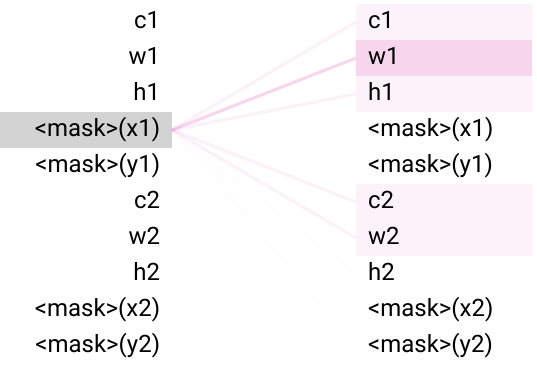}
        \caption{head 2-6}
    \end{subfigure}%
    ~ 
    \begin{subfigure}[t]{0.23\textwidth}
        \centering
        \includegraphics[width=0.95\textwidth]{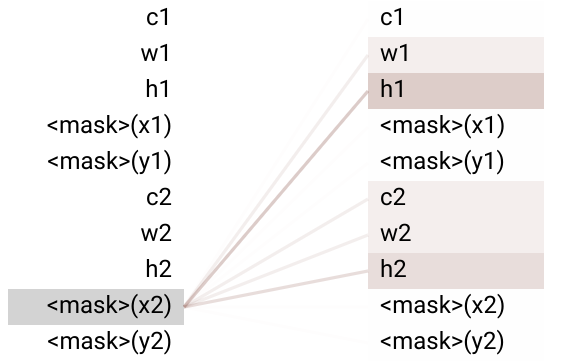}
        \caption{head 2-5}
    \end{subfigure}
    ~
    \begin{subfigure}[t]{0.23\textwidth}
        \centering
        \includegraphics[width=0.95\textwidth]{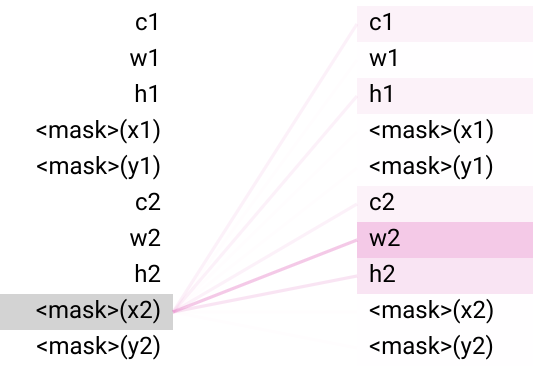}
        \caption{head 3-6}
    \end{subfigure}
    ~
    \begin{subfigure}[t]{0.23\textwidth}
        \centering
        \includegraphics[width=0.95\textwidth]{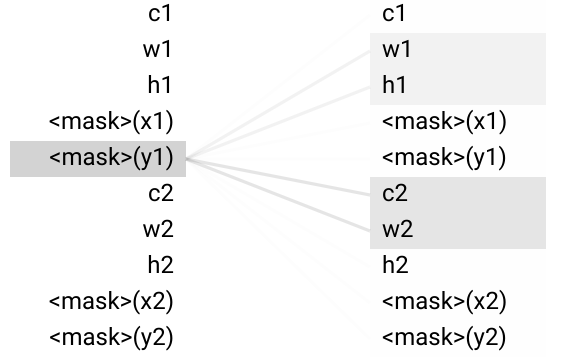}
        \caption{head 3-7}
    \end{subfigure}
    \caption{Additional examples of attention heads exhibiting the patterns for masked tokens. The darkness of a line indicates the strength of the attention weight (some attention weights are so low they are invisible).  We use $\langle$layer$\rangle$-$\langle$head number$\rangle$ to denote a particular attention head.}
    \label{fig:head-vis-more}
\end{figure*}

Patterns for other heads at different layers are listed in Fig.~\ref{fig:head-vis-more}. We could find that for masked $x$ position (head 1-1 and head 2-6, \etc), their heads will attend to width information of various objects for accurate prediction. And similar findings could be found for other heads.

\subsection{Failure Cases}\label{sec:failure_cases}
Some undesired conditional generation results are shown in Fig~\ref{fig:failure}. Similar to other layout generation models, there are some overlaps between objects in some generation results. Furthermore, some generated samples are largely different from the real layouts with low visual quality. For example, in the second sample on the Magazine, the alignment of the generated sample is worse than its corresponding real layout. We will explore these directions in the future work.

\subsection{Iterative Refinement Process} 
\begin{figure}[!ht]
	\centering
	\includegraphics[width=0.95\textwidth]{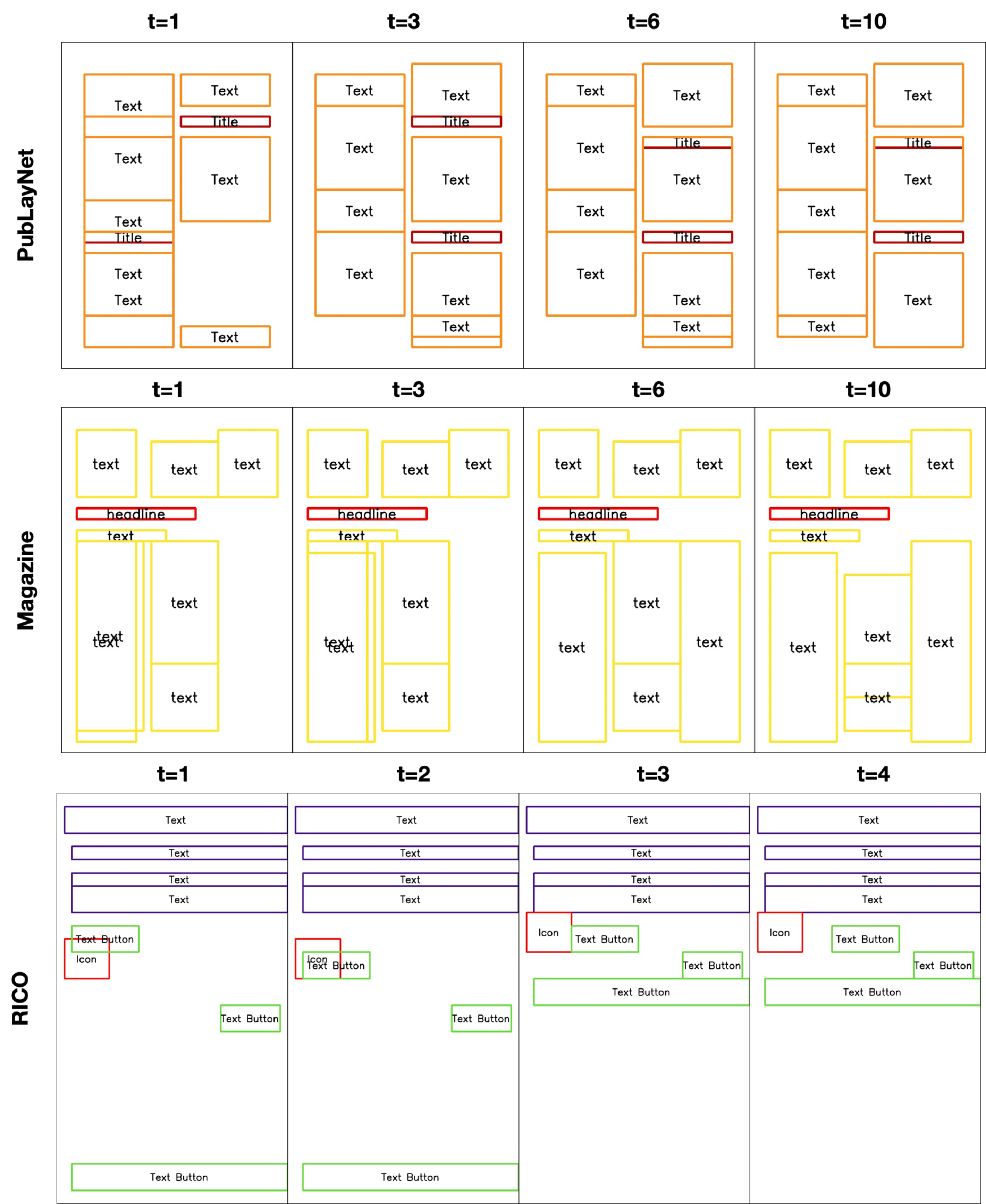}
	\caption{Layouts refinement process. Layouts generated at different iterations ($t$) are shown on three datasets.}
	\label{fig:refine}
\end{figure}

To understand the process of our iterative refinement algorithm, we explore the performance of models with various iterations. Quantitatively, IOU and Overlap metrics, where the lower, the better, are plotted in Fig.\ref{fig:refine_scores} along with the number of iterations for refinement. With more iterations, the quality metrics are getting improved and stable.
We also show samples of generated layouts at different number of iterations in Fig.~\ref{fig:refine}. At the first iteration, there are severe overlaps between objects, showing the difficulty to yield high-quality layouts with just one pass. However, after iteratively refining low-confident attributes, the layouts become more realistic.

\end{document}